\def\year{2022}\relax
\documentclass[letterpaper]{article} %
\usepackage{aaai22}  %
\usepackage{times}  %
\usepackage{helvet}  %
\usepackage{courier}  %
\usepackage[hyphens]{url}  %
\usepackage{graphicx} %
\usepackage{natbib}  %
\usepackage{caption} %
\DeclareCaptionStyle{ruled}{labelfont=normalfont,labelsep=colon,strut=off} %
\usepackage{algorithm}
\usepackage{algorithmic}
\usepackage[algo2e,ruled,vlined,noend]{algorithm2e}
\usepackage{booktabs}       %

\usepackage{subcaption}
\usepackage{mathtools}
\usepackage{amsmath}
\usepackage{amsfonts}
\usepackage{bbm}
\usepackage{thmtools} 
\usepackage{thm-restate}

\newtheorem{definition}{Definition}[section]

\newtheorem{lemma}{Lemma}
\newcommand{\alg}{$\mathcal{A}lg$}

\newcommand{\J}{\bar{J}}
\usepackage{newfloat}
\usepackage{listings}
\floatstyle{ruled}
\newfloat{listing}{tb}{lst}{}
\floatname{listing}{Listing}
\title{Constraint Sampling Reinforcement Learning: Incorporating Expertise For Faster Learning}
\author {
    Tong Mu\textsuperscript{\rm 1},
    Georgios Theocharous \textsuperscript{\rm 2},
    David Arbour\textsuperscript{\rm 2},
    Emma Brunskill \textsuperscript{\rm 1}
}
\affiliations {
    \textsuperscript{\rm 1} Stanford University, \{tongm, ebrun\} @ cs.stanford.edu\\
    \textsuperscript{\rm 2} Adobe Research, \{theochar, arbour\} @ adobe.com
}
\begin{document}

\maketitle

\begin{abstract}
Online reinforcement learning (RL) algorithms are often difficult to deploy in complex human-facing applications as they may learn slowly and have poor early performance. To address this, we introduce a practical algorithm for incorporating human insight to speed learning. Our algorithm, Constraint Sampling Reinforcement Learning (CSRL), incorporates prior domain knowledge as constraints/restrictions on the RL policy. It takes in multiple potential policy constraints to maintain robustness to misspecification of individual constraints while leveraging helpful ones to learn quickly. Given a base RL learning algorithm (ex. UCRL, DQN, Rainbow) we propose an upper confidence with elimination scheme that leverages the relationship between the constraints, and their observed performance, to adaptively switch among them. We instantiate our algorithm with DQN-type algorithms and UCRL as base algorithms, and evaluate our algorithm in four environments, including three simulators based on real data: recommendations, educational activity sequencing, and HIV treatment sequencing. In all cases, CSRL learns a good policy faster than baselines.
\end{abstract}

\section{Introduction}
Online reinforcement Learning (RL) algorithms have the large potential for improving real world systems with sequential decisions such as recommendation systems\cite{theocharous2020reinforcement} or intelligent tutoring systems~\cite{bassen2020reinforcement}. 
Such domains often have large or infinite state spaces, and existing RL methods that can scale to these settings frequently require a prohibitively large  amount of interaction data to learn a good policy. Incorporating human expert knowledge can accelerate learning, such as through expert demonstrations~\cite{wu2019imitation,arora2021survey,hussein2017imitation,taylor2011integrating} or having them provide properties the optimal policy is guaranteed to satisfy, such as a specific function class constraint~\cite{ijspeert2002movement,tamosiunaite2011learning,buchli2011learning, kober2013reinforcement}. However, such approaches also run the risk of reducing the system performance when the provided information is misleading or suboptimal. For example, in recommendation systems, 
prior Human Computer Interaction (HCI) literature has found that diversity of recommendations across time is  important~\cite{nilashi2016recommendation,bradley2001improving,komiak2006effects, chen2019serendipity}. But confidently concluding this finding will generalize to a new system, and translating this high level knowledge into a concrete constraint for the policy class is subtle (does 'diversity' mean 3 or 4  different item categories? Within the last 5 or the last 10 recommendations?). Choosing the wrong formulation can significantly impact the resulting reward. 

An alternate approach is to allow a human domain expert to provide many different formulations, and use model or algorithm selection to automatically learn to choose the most effective (e.g.~\citep{lee2021online,laroche2018reinforcement}. However, much of the work on selection focuses on theoretical results and cannot be used with popular deep RL algorithms that may be of more practical use~\cite{lee2021online}. Perhaps most closely related to our work is the work of \citet{laroche2018reinforcement}, which uses an upper confidence bound bandit approach to learn to select among a set of reinforcement learners. This work was constructed for optimizing across learning hyperparameters (such as learning rate or model size) and, like many upper confidence bound approaches, relies on tuning the optimism parameter used, which is often hard to do in advance of deployment.

In this work we instead focus on leveraging human insight over the domain to speed RL through weak labels on the policy. We propose an algorithm, Constraint Sampling Reinforcement Learning (CSRL)\footnote{Code: \url{https://github.com/StanfordAI4HI/CSRL}} which performs adaptive selection and elimination over a set of different policy constraints. Our selection algorithm optionally use these constraints to learn quickly, and to distinguish this from the safety constraints used in safe RL, we will also refer to them as \textit{policy restrictions}. A policy constraint or restriction limits the available actions in a state and can speed learning by potentially reducing the exploration of suboptimal actions. 
Our method performs algorithm selection over a set of RL learners, each using a different policy restriction. For example, in a recommendation system, one restriction might specify that 3 items should be from unique categories in the past 5 items shown, and another could require at least 2 unique items in the past 10, but only for users who have used the system for more than 6 months. A third RL learner could use the unrestricted policy class, which does not limit the available actions in any state. 
Note that other areas of machine learning have significant reductions in cost by allowing people to provide weak supervision through labeling functions that may be imperfect(e.g.~\cite{ratner2017snorkel}). 
At a high level, we apply a similar idea here for reinforcement learning, allowing people to provide weak, potentially imperfect policy restrictions to be used by different RL learners.
Our algorithm then performs adaptive selection over the set using optimism under uncertainty over the potential performance of the RL learners, each of which is operating with a different set of restrictions. A second technical innovation comes from noting that tuning the optimism parameter of the selection mechanism in advance can be infeasible, and a parameter either too high or too low can slow learning. Instead, we introduce a simple heuristic which uses  the amount of model change to estimate the convergence of each RL learner and use this to eliminate RL learners with low performance. This allows us to achieve good performance much faster than through optimistic adaptive selection alone.

These simple ideas lead to substantial empirical improvements in the diverse range of settings we consider, which include simulators created with real data in movie recommendations, tutoring system problem selection, and HIV treatment, as well as the Atari lunar lander simulator. We conduct a careful experimental analysis to illustrate the benefits of our additional change-based learner elimination and the robustness to the inclusion of poor constraints. Our approach is simple and can be used with a wide variety of base reinforcement learners, and may make RL more feasible for a wider set of important domains. 

\section{Setting}
 A Markov Decision Process (MDP)~\cite{bellman1957markovian} is described as a tuple $(\mathcal{S}, \mathcal{A}, P, R)$ where $\mathcal{S}$ is the set of states and $\mathcal{A}$ is the action set. The transition dynamics, $P(s'|s, a)$ defines the probability distribution of next states $s'$ after taking action $a$ in state $s$, and $R(r|s,a)$ defines the distribution of rewards $r$. We assume the action space is discrete (however the state space can be either discrete or continuous) and rewards are bounded $|r| < R_{max}$. We consider the episodic, finite horizon setting where the length of each episode is less than or equal to the maximum horizon length $H$.
A policy, denoted as $\pi$ is a potentially stochastic mapping from states to actions, where $\pi(a|s)$ defines the probability of taking action $a$ in state $s$.
A trajectory, $\tau \coloneqq (s_0, a_0, r_0, s_1, a_1, r_1, ...)$, is defined as the sequence of states, actions, and rewards in an episode. The state-action value of a policy, $Q_{\gamma}^{\pi}(s, a)$, is the expected discounted sum of rewards of starting in a state $s$, taking action $a$ and then following the policy: $Q_{\gamma}^{\pi}(s, a) \coloneqq \mathbb{E}_{\tau \sim \pi} [\sum_{t = 0}^{h}\gamma^t r_t|s_0 = s, a_0 = a]$), where $\gamma \in [0, 1]$ is the discount factor. The value function is the expected discounted sum of rewards obtained starting in state $s$ and following policy $\pi$: 
$V_{\gamma}^{\pi}(s) \coloneqq \mathbb{E}_{\tau \sim \pi} [\sum_{t = 0}^{h}\gamma^t r_t|s_0 = s]$
The optimal policy, denoted $\pi^*$, is the policy that maximizes $V$: $\pi^* = \arg \max_\pi V_{\gamma}^{\pi}$. 

\section{Algorithm}

We first briefly present the aim and overview of our algorithm, Constraint Sampling Reinforcement Learning (CSRL), before going  into detail. The goal of our work is to provide a simple method for leveraging (potentially weak) human domain knowledge for faster learning without sacrificing final performance. CSRL takes as input a set of different candidate policy \textit{constraints} (we also refer to as \textit{restrictions} to distinguish our goal from that of the Safe RL work which uses constraints), each of which is used to define a RL learner that must follow the restriction while learning. Some or all of the provided restrictions may disallow the (unknown) optimal policy and the unrestricted option may be included as a candidate.
CSRL then uses an upper confidence bandit (UCB)~\cite{auer2002using} method that considers the prior performance of the learners to adaptively select the next learner to use to make decisions for the episode. In practice such optimism based approaches often require careful tuning to achieve good performance which can be infeasible in new domains lacking prior data. To increase robustness to this hyperparameter and speed learning, our CSRL method introduces a simple heuristic that tracks the model parameter changes in the RL learner for hypothesizing when a learner has converged, and eliminates those that may be misspecified or have low performance. 

We now introduce some notation. In this paper we use the following definition of constraint/restriction: 
\begin{definition}[Constraint/Restriction] A constraint (or restriction) $C$ is a function that maps each state to a set of allowable actions, $C(s) = \{a_i,a_k,\ldots \}$.\footnote{Note this in is contrast to the Constrained MDP (CMDP) framework~\cite{altman1999constrained}, which has a different focus on costs and budgets.}
\end{definition}
Given a set of $K$ restrictions $\mathcal{C}$, let $C_k$ denote the $k^{th}$ restriction.
We say a policy $\pi$ \textit{satisfies} a restriction $C_k$ if for every state, $\pi$ only takes actions allowed by $C_k$:  $\forall (s,a)$ $\pi(a|s) > 0$ only if $a \in C_k(s) $. 
\begin{definition}[Restricted Policy Set] We denote the policy set of restriction $C_k$ as $\Pi_k$ and define it as the set of all policies $\pi$ that satisfy $C_k$:
$$\Pi_k = \{ \pi : \forall(s,a) \pi(a|s) > 0 \to a \in C_k(s) \}$$ %
\end{definition}

\begin{definition}[Restricted RL Learner] Given a restriction $C_k$, and a base RL learning algorithm that can learn under constraints (such as DQN) we instantiate a \textbf{restricted reinforcement learner}, denoted $l_k$. $l_k$ is restricted to executing and optimizing over policies in $\Pi_k$
\end{definition}

 We assume each restriction in the set is unique and we define the subset property between restrictions:
\begin{definition}[Subset/More Restricted]
Restriction $C_k$ is a \textit{subset} of restriction $C_j$ if every action allowed in $C_k$ is also allowed in $C_j$: $\forall s \ C_k(s) \in C_j(s) $. In this case, we will also refer to $C_k$ as more restricted than $C_j$ and define the $<$ operator: $C_k < C_j$. We also apply this notation to describe the corresponding policy sets and RL learners.
\end{definition}

\noindent \textbf{Note on specifying restrictions:} Note that while policy restrictions are defined by the state-action pairs allowed, they often do not need to be specified by humans at that level of granularity. For example, a human expert might specify that students should only be given practice problems involving at most one new skill. As long as the state and action space has features representing the skills involved, it is easy to programmatically translate this high level specification to the desired constraint without requiring the human expert to enumerate all state-action pairs.

Our algorithm, CSRL (Algorithm~\ref{alg:CSRL}), takes as input a base RL algorithm, \alg,  a set of potential restrictions $\mathcal{C}$ and a confidence bound function $B(h, n)$.
CSRL starts by initializing $|\mathcal{C}|$ RL learners, each which use the input base algorithm \alg, along with one of the restrictions $C_i \in \mathcal{C}$. Learner $l_i$ will only chose actions and learn over policies that satisfy $C_i$. Let $\mathcal{L}$ denote the set of active learners and initially set  $\mathcal{L}=\{1,\ldots,|\mathcal{C}| \}$. 
Each episode proceeds in 3 steps. First CSRL chooses a RL learner $l_i$ in the active set $\mathcal{L}$ to select actions in this episode. Second, CSRL gathers a trajectory of states, actions, and rewards using $l_i$. This data is used both to update the learner $l_i$ as well as all the relevant other learners: sharing data speeds learning across all the RL learners. Third, a heuristic is used to potentially eliminate $l_i$ from $\mathcal{L}$. The next episode then starts. 

We now describe step 1, RL learner selection. We use UCB to select the RL learner with the maximum estimated upper bound on the potential return in the active set. The UCB mechanism uses the average prior returns observed during past executions of RL learner $l$, denoted $\hat{\mu}_{l}$, and the input confidence bound function.  

\begin{equation}\label{eq:UCB_Const}
    k = \arg \max_{l \in \mathcal{L}} \left( \hat{\mu}_{l} + B(h, n_l)\right)
\end{equation}
There are two important things to note about this learner selection strategy. First, CSRL does not use any RL learners' internal estimates of their own performance or potential value function. This allows us to leverage very general base algorithms $Alg$ without requiring that they accurately estimate their own value functions. Instead CSRL relies only on the observed returns from executing each learner, treating them as one arm in a multi-armed bandit. Second, note that the estimated upper bound for a given learner $l$ in Equation~\ref{eq:UCB_Const} will generally not be a true upper confidence bound on the performance of the RL algorithm. This is because the UCB multi-armed bandit algorithm assumes that the stochastic returns of individual arms are unknown, but stationary. In contrast, in our setting, arms are RL learners whose policies are actively changing. Fortunately, prior related work has successfully used UCB to select across arms with non-stationary returns in an effective way: this is the basis of the impactful upper confidence trees, a Monte Carlo Tree Search method that prioritizes action expansions in the tree according to upper bounds on the non-stationary returns of downstream action decisions~\cite{shah2020non}. It has also been used in related work on online RL~\cite{laroche2018reinforcement}, and we will demonstrate it can both be empirically effective in our setting, and, under mild assumptions, still guarantee good asymptotic performance. 

After gathering data using the selected RL learner, this data is provided to all RL learners to optionally update their algorithm\footnote{A reader may wonder if providing such off policy data is always useful or how best to incorporate it.  In the particular base RL algorithms we use, it is straightforward to incorporate the observed trajectories into experience replay or in an estimate of dynamics and reward models, but more sophisticated importance sampling methods could be used for policy gradient based methods.}. 
Then the third step is to eliminate the potential learner associated with the chosen constraint. The elimination heuristic checks if a RL learner $l_i$'s value or state-action value  has stabilized, and if its average empirical performance is lower than another RL learner $l_j$, we will eliminate $l_i$ if a less constrained learner $l_k$ is in the active set. We now state this more formally.

When RL learner $l_k$ is used for the $n^{th}$ time, and generates trajectory $\tau_n$, let $\delta^n_{k}$ represent the change in the value function. For example, in tabular domains, we can measure the average absolute difference in the state-action values
\begin{equation}\label{eq:loss}
    \delta^n_{k} =  \frac{||Q_k^{n-1} - Q_k^n||_{1}}{|\mathcal{S}||\mathcal{A}|},
\end{equation}
where $Q_k^n$ represents the state-action value function after updating using the data just gathered. In value-based deep RL we can use the loss function over the observed trajectory as an approximate estimate of the amount change\footnote{Note one could use other measures of the change in the RL learner, including differences in the neural network parameters, or changes in predictions of the value of states.}
\begin{gather}\label{eq:dqnloss}
    \delta^n_{k} = \sum_{\tau^n} (r_t + \max_{a \in C_k(s_{t+1})}Q^n_{k}(s_{t+1}, a)) 
    - Q^n_{k}(s_t, a_t).
\end{gather}

For a RL learner to be potentially eliminated, the change, $\delta^{n}_{k}$, must be below a threshold $T_l$ for at least $T_n$ successive steps, suggesting that the underlying learning process is likely to have stabilized. If this condition is satisfied for a RL learner $l_i$ the meta-leaner first checks there exists a less constrained learner $l_k$ in $\mathcal{L}$. If such a $l_k$ exists and at least one other learner $l_j$ in $\mathcal{L}$ has higher average performance: $\hat{\mu}_i < \hat{\mu}_j$, then $l_i$ is removed. 
See Algorithms~\ref{alg:CSRL} and~\ref{alg:elim} for pseudocode. We give examples of instantiations of CSRL with base learners of UCRL and DDQN in section~\ref{sec:specifics} below and we will release all our code on Github. 

\noindent \textbf{Elimination Mechanism Intuition: } 
Recall that CSRL should try to select the most restricted learner that is compatible with the optimal policy, since learning with a reduced  action space will often speed learning and performance. To increase robustness, the RL learner selection UCB strategy relies on minimal information about the internal algorithms. However, UCB often can be conservative in how quickly it
primarily selects high performing arms (in our case, RL learners) rather than lower reward arms (in our case, learners with restrictions incompatible with the optimal policy). 
Consider if one can identify when learner $l_i$ has converged %
and can correctly evaluate its performance through the running average $\mu_i$. %
If there exists $l_j$ with higher average returns $\mu_j$, restriction $C_i$ is likely to not include the optimal policy and $l_i$ can be removed from the set of potential RL learners. 
Our method uses a proxy heuristic for convergence and sample estimates for returns which can be noisy. Therefore it may incorrectly conclude a learner $l_i$ has converged and/or has suboptimal returns. %
To ensure that we preserve a learner that admits the optimal policy, we only eliminate a learner $l_i$, and its constraint $C_i$, if it is a subset of at least one other active constraint $C_j$. Therefore the the set of policies that satisfy $C_i$ will continue to exist in the active set, even when $C_i$ is eliminated.
We now provide some basic soundness to this proposed approach, before describing instantiations of CSRL with particular base learners, and demonstrating its performance empirically.

\begin{figure}[t]
    \begin{minipage}{0.46\textwidth}

\begin{algorithm}[H]
    \caption{CSRL}\label{alg:CSRL}
    \DontPrintSemicolon
    \SetAlgoLined
    \SetKwFunction{I}{Initialize}
    \SetKwFunction{O}{ConstraintedPolicy}
    \SetKwFunction{GT}{GenerateTrajectory}
    \SetKwFunction{U}{UpdateLearners}
    \SetKwFunction{E}{Eliminate}
    \SetKwFunction{PC}{$\mathcal{M}$Change}
    \textbf{Inputs:} $\mathcal{A}lg$, $\mathcal{C}$, $\mathcal{Z}$, $B(h,n)$, $T_l$, $T_n$\\
    \textbf{Initialize:} $\mathcal{L} \leftarrow$ Create Restricted Learners from $\mathcal{C}$\\
     $D_{k} = [\ ] \ \forall C_k \in \mathcal{C}$\tcp*{Model Change Amounts}
    
    \For{Episode $h = 1, 2, ...$ }{
    $l_k \leftarrow$ select RL learner [Eqn.~\ref{eq:UCB_Const}]\\
    $\tau_h \leftarrow$ \GT($l_k$)\\ %
    $R_h = \sum_{t=1}^{len(\tau_h)} r_t$\\
    $n_{k}=n_k + 1$\\
    $\hat{\mu}_{k} = \frac{(n_{k}-1)\hat{\mu}_k + R_h}{n_k}$\\
     \U($\mathcal{L}, \tau_h$)\\
    $\delta_h \leftarrow$ Calculate Change of $l_k$ [ eq.~\ref{eq:loss} or 3]\\
    $D_{k} \leftarrow [D_{k} , \delta_h]$\\
    \If{ \E($l_k, D_{k}$)}{
    $\mathcal{L} \leftarrow \mathcal{L} \setminus {l_k}$
    }
    }
\end{algorithm}
\end{minipage}
\begin{minipage}{0.45\textwidth}
\begin{algorithm}[H]
    \caption{Eliminate (Eliminate Learner)}\label{alg:elim}
    \DontPrintSemicolon
    \SetAlgoLined
    \textbf{Inputs:} ${l_k}, D_{k}$
    
    \If{$\exists n \in \{n_k-T_n:n_k\}$ such that $D_k(n) > T_l$}{
    Return \textit{False} (Don't Eliminate)
    }
    \If{Exists $l_i, l_j \in \mathcal{L}: C_i > C_k \text{ and } \hat{\mu}_{j}> \hat{\mu}_{k}$}{
    Return \textit{True} (Eliminate)
    }
    Return \textit{False}

\end{algorithm}
\end{minipage}
\end{figure}

\subsection{Brief Theoretical Discussion} \label{sec:guarantee}
We briefly provide a guarantee that at least one learner whose policy set contains the optimal policy will be taken more than all other learners under mild assumptions.
\begin{restatable}[Model Parameter Convergence]{assump}{assumpmpc}
\label{thm:assumpmmpc}
Let $M_{k,n}$ represent the model parameters of the learner $l_k$ after the $n^{th}$ update. In every run of CSRL, the model parameters of every learner converge:
$\lim_{n \rightarrow \infty} \mathcal{M}_{k, n} \rightarrow \mathcal{M}_k \ for \ all \ C_k \in \mathcal{C}.$ 
\end{restatable}

Let $\pi_{\mathcal{M}_k}$ denote the policy corresponding to the converged model parameters $\mathcal{M}_k$ of RL learner $l_k$. Let $\mathbb{E}_{s_0}[V^{\pi_{\mathcal{M}_k}}(s_0)] = \mu_k$, $\mu^* = \max_k \mu_k$ and $\pi^*$ denote the policy that achieves $\mu^*$. Note that $\pi^*$ is defined as the policy with the highest return across all learners in the set.   Without loss of generality assume $\pi^*$ is unique. We refer to the set of constraints compatible with $\pi^*$ as the set of optimal constraints and denote this set as $\mathcal{C}^*$ with a corresponding indices set $\mathcal{K}^*$. Then require:
\begin{restatable}
[Convergence to Optimal]{assump}{assumpc}
\label{thm:assumpc}
Given a specific run of CSRL, let every learner in $\mathcal{K}^*$ converge to the $\mu^*$ of the run:
$\mu_k = \mu^* \ for \ all \ k \in \mathcal{K}^*$ 
\end{restatable}

 We now show that at least one RL learner in $\mathcal{C}^*$ will be chosen more than all suboptimal learners asymptotically. 

\begin{restatable}
[]{thm}{thmr}
\label{thm:r} Assume Assumptions 1 and 2 hold. Also assume as input a confidence bound $B(h,n)$  of the form $\frac{z(h)}{n^\eta}$ with $0 < \eta < \frac{1}{2}$ and $z(h)$ satisfying the following two conditions: (i) $z(h)$ is non-decreasing and (ii) $O(z(h)^{1/\eta}) < O(h)$. Let $T_k(h)$ be the number of times RL learner $l_k$ has been selected at episode $h$. Then for at least one $k^* \in \mathcal{K}^*$, $T_{k^*}(h) > T_k(h)$ for all $k \notin \mathcal{K}^*$ as $h \rightarrow \infty$
\end{restatable}
The proof of theorem 1 is provided in the appendix. 

When the base algorithm has convergence guarantees, such as UCRL, we can additionally provide guarantees on the rate of convergence. We provide these rates and a discussion of the UCRL case in the appendix: our analysis drawns upon the analysis of convergence rates for Monte Carlo Tree Search from Shah et al.~\cite{shah2020non}. 

\subsection{Algorithm Instantiations}
\label{sec:specifics}
We discuss specific instantiations of CSRL with various base RL algorithms.

\noindent \textbf{CSRL-UCRL}. UCRL~\cite{auer2009near} is an influential strategic RL algorithm with good regret bounds this is based on optimism under uncertainty for tabular settings. It is simple to incorporate action constraints during the value iteration steps:
\begin{equation*}
    V^{t+1}_h(s) \leftarrow \max_{a \in C_h(s)} (\tilde{R}(s,a) + \sum_{s'} \tilde{P}(s'|s,a) \gamma V^t_h(s'))
\end{equation*}
All observed $(s,a,r)$ tuples are used to update an estimated model of transitions and rewards that is shared across all RL learners. Equation~\ref{eq:UCB_Const} is used to track convergence in the estimated value function.

\noindent \textbf{CSRL-DQN, CSRL-DDQN and CSRL-Rainbow}
Deep reinforcement learning has shown powerful results across a wide variety of complex domains. 
Our CSRL-DQN, CSRL-DDQN, and CSRL-Rainbow implementation uses a separate DQN~\cite{mnih2015human}, DDQN~\cite{van2015deep}, or Rainbow~\cite{hessel2018rainbow} learner for each restriction\footnote{ We explored sharing some model weights but found that resulted in worse performance}. We used epsilon greedy exploration with epsilon decay. Experience is shared across learners in the form of a shared replay buffer. The \texttt{UpdateLearners} function places the tuples from the most recent trajectory in the shared replay buffer. Each learner $l_k$ is then updated, using only samples from the buffer that satisfy the associated restriction $C_k$. 
The learner $l_k$ is updated using 
a constrained Q loss~\cite{kalweit2020deep} (see Equation 3).

\section{Experiments}
We briefly introduce the evaluation environments and the constraints used and then discuss our results. Due to space constraints, we defer detailed descriptions  of the environments and constraint constructions to the appendix. 

\begin{figure}[t]
    \begin{subfigure}[b]{0.15\textwidth}
     \centering

    \includegraphics[width=\textwidth]{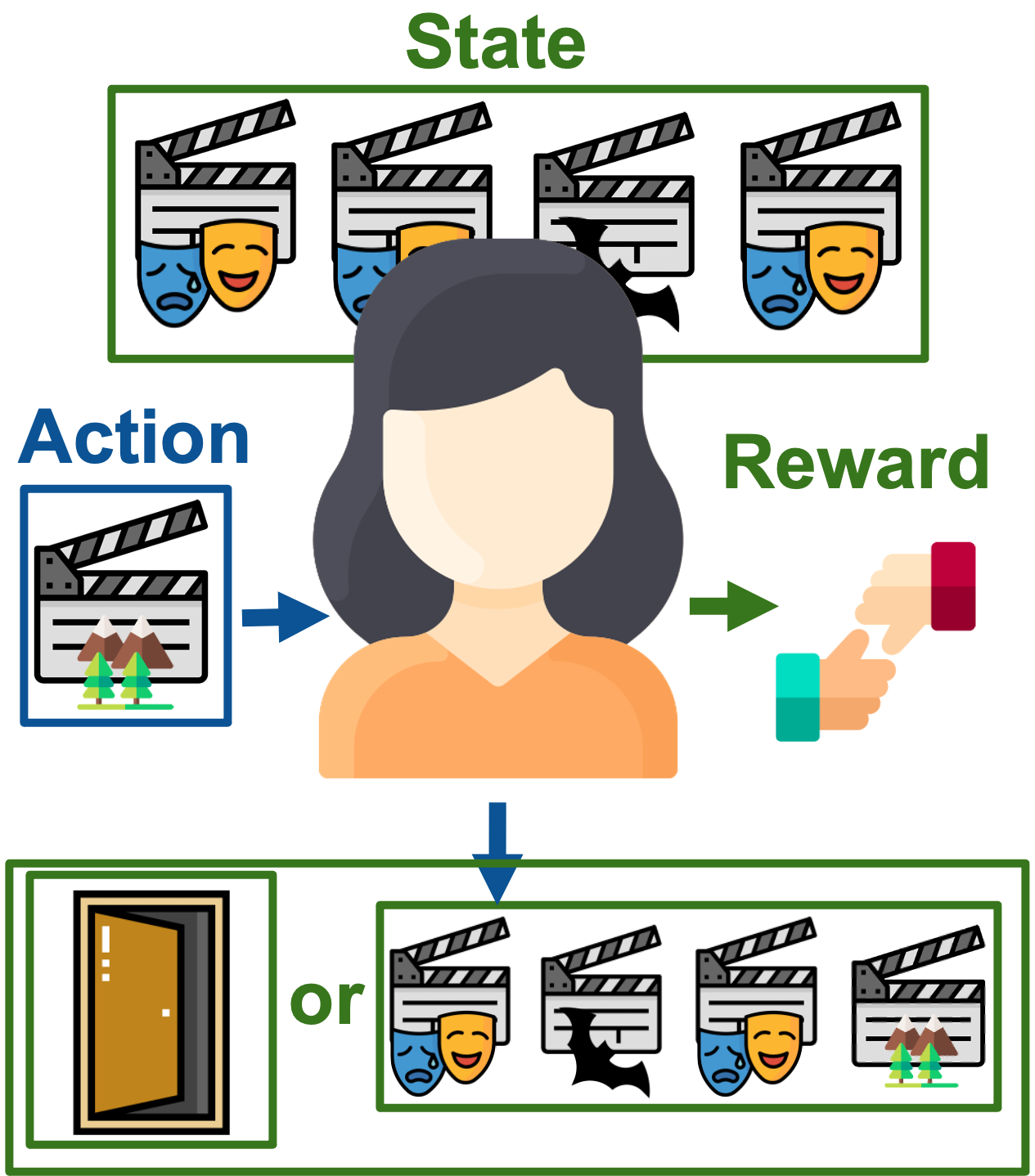}
    \caption{}
         \label{fig:rs_env}
     \end{subfigure}
     \begin{subfigure}[b]{0.15\textwidth}
     \centering
    \includegraphics[width=\textwidth]{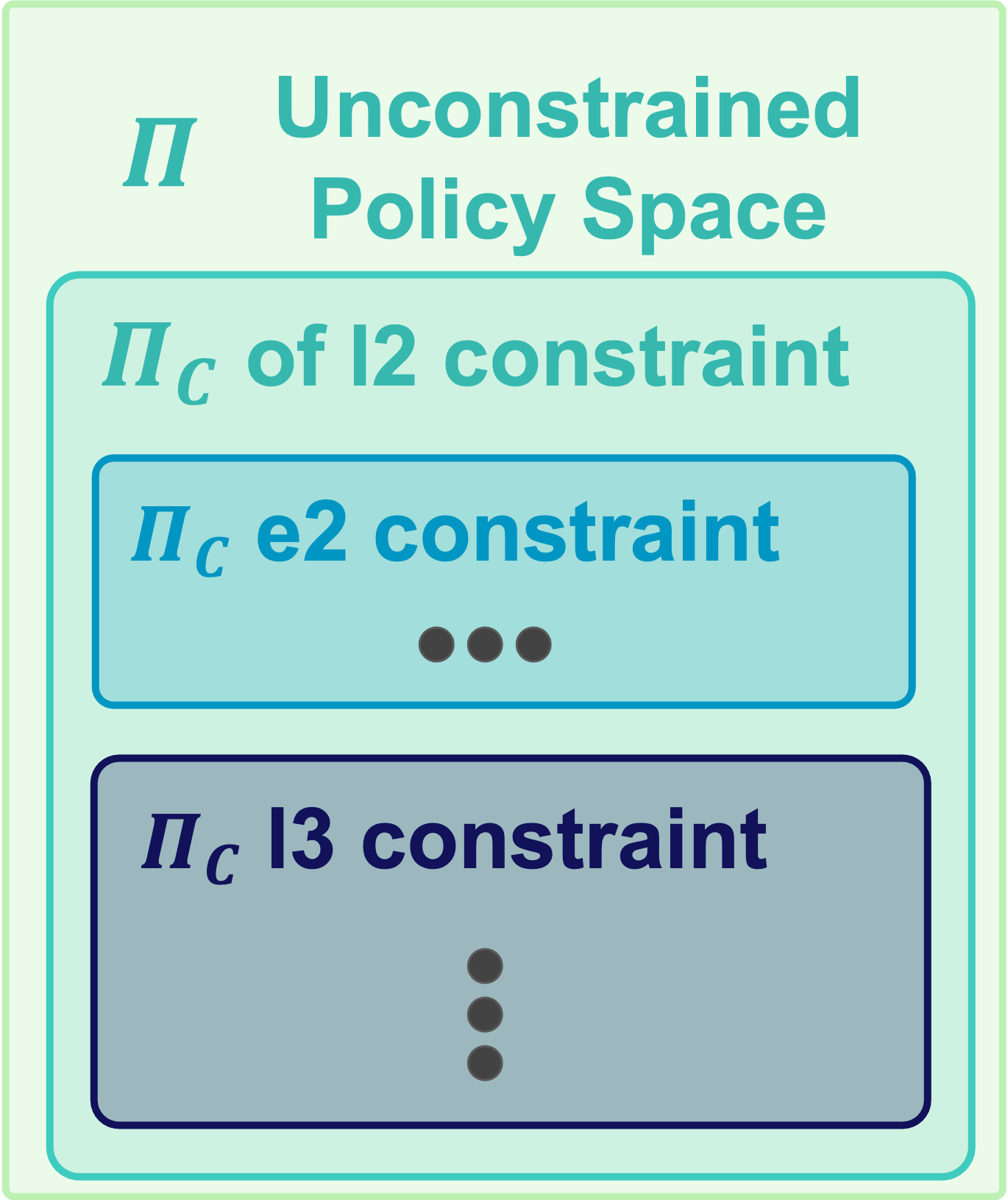}
    \caption{}
         \label{fig:rs_cs}
     \end{subfigure}
     \begin{subfigure}[b]{0.15\textwidth}
     \centering
     \includegraphics[width=\textwidth]{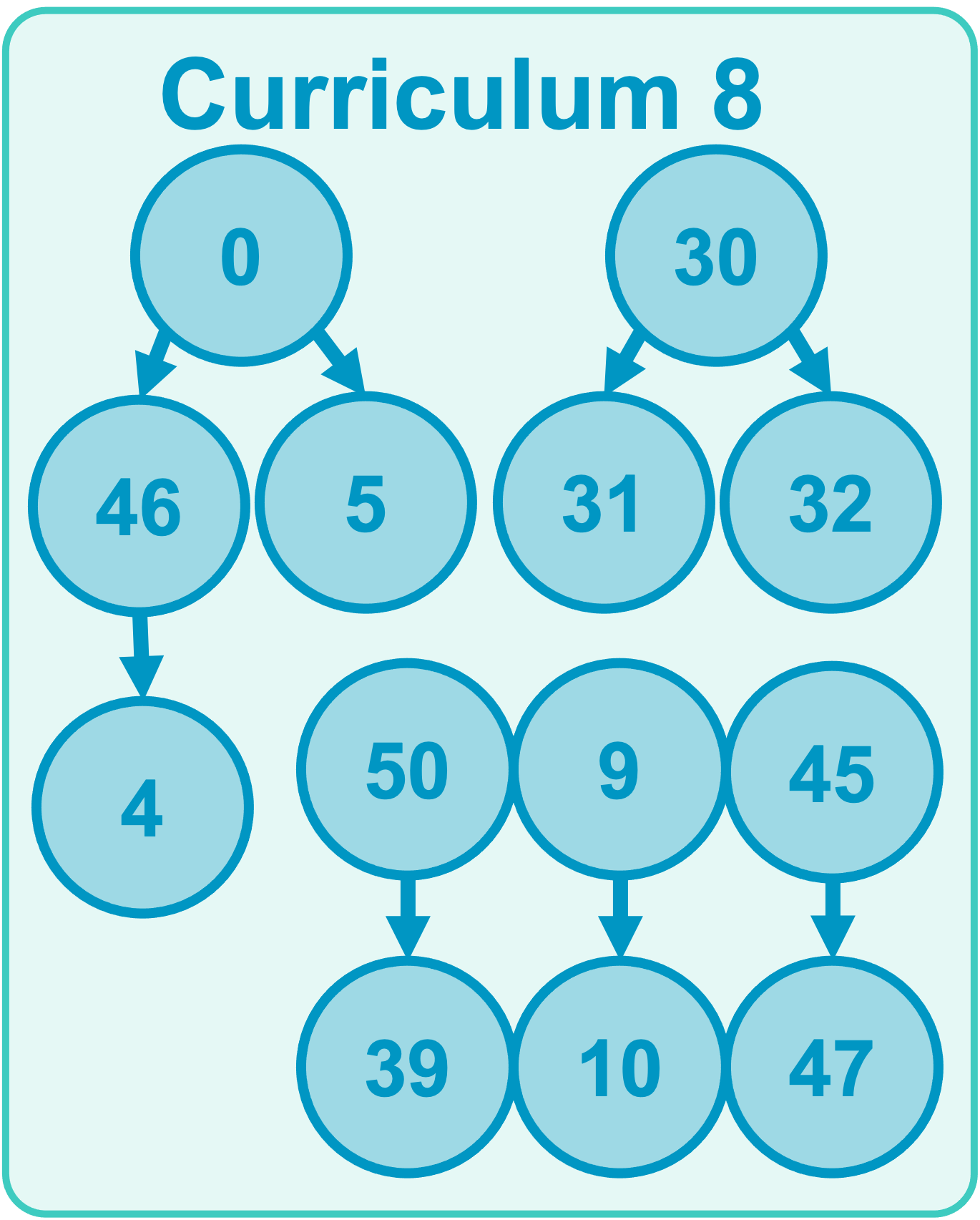}
    \caption{}
         \label{fig:C8}
     \end{subfigure}
\caption{(a) An image of the recommendation system environment (RS env.) (b) A visualization of the policy space for the RS env. For example, every policy allowed under the \textit{exactly 2 variability} constraint ('e2') is also allowed under the \textit{at least 2 variability} ('l2') constraint. (c) An example curriculum graph in the education domain. Each directed edge indicates the source node is a prerequisite of the sink node.}
\label{fig:envs}
\end{figure}

\subsection{Environments and Constraints}

\textbf{Recommendation System Environment}
The movie recommendations (See Fig~\ref{fig:rs_env} for illustration) environment is a slighlty modified environment from prior work~\cite{warlop2018fighting} fit with the Movielens 100K dataset~\cite{harper2015movielens} which contains 100K user-movie ratings across 1000 users with $|\mathcal{A}| = 5$ and $|\mathcal{S}| = 625$.
Each action is a movie genre and the state space encodes recently seen movies.
The rewards correspond to the ratings of the recommended movies. The episode length is random and each state-action pair has some probability of leading to the terminal state. 

Following prior work that suggests diversity is correlated with system usage~\cite{nilashi2016recommendation}, we design a set of 12 constraints using a \textit{variability} factor, which we define as the number of unique actions in recent history. Our constraints require the policy to maintain a certain level of variability.
A partial visualization of the structure between some constraints in the set is given in Figure ~\ref{fig:rs_cs}. Because our state space contains the history of past actions, these high level constraint specification are easily translated programmatically.

\textbf{Education: Deep Knowledge Tracing Student.}
Our new educational activities sequencing environment uses a popular student learning model, the Deep Knowledge Tracing (DKT) model~\cite{piech2015deep}, to simulate students. The model is trained with the ASSISTment 2009-2010~\cite{feng2009addressing} dataset of algebra learning, containing data from 
3274 students
over 407K student-problem interactions.
Each action ($|\mathcal{A}| = 51$) corresponds to presenting a different problem to the student. 
The horizon is length $H = 100$. The state space is a continuous $\mathbb{R}^{58}$ and encodes the current proficiency (the predicted probability of mastery by the DKT model) on each problem and the binary encoded timestep.
The reward corresponds to a signal of mastery and is $1$ when the proficiency of a problem first exceeds a threshold $m_t = 0.85$ and $0$ otherwise.

In education, curriculum or prerequisite graphs are common; however, setting the correct granularity to model knowledge can be difficult. We create a constraint set consisting of different graphs: Figure~\ref{fig:C8} shows an example using automatic curriculum generation methods that requires hand specifying a hyperparameter to adjust the number of edges~\citep{piech2015deep}.  
We construct a constraint set containing 13 different graphs. Given a graph, we only allow unmastered problems that have all prerequisites mastered (this information is encoded in the state space) to be selected.

\textbf{HIV treatment Simulator}
The HIV treatment simulator~\cite{adams2004dynamic,ernst2006clinical} simulates a patient's response to different types of treatments. The action space is discrete with size 4 and represents various treatment actions. The state space is continuous $\mathbb{R}^6$, where each dimension represents a marker of the patient's health. Each episode is 200 timesteps and the reward encourages patients to transition to and maintain a healthy state while penalizing drug related actions. 

We created a simulator for multiple heterogeneous patient groups by perturbing the internal hidden parameters of the system following \citet{killian2017robust}. %
We learn an optimal decision policy for 3 different groups, and then use the known optimal policies as constraints to learn in a new group which may or may not be similar to a group with a known policy. We create a constraint set of 7 constraints. 

\textbf{Lunar Lander} The Lunar Lander environment from Open AI Gym~\cite{brockman2016openai} simulates landing an aircraft. The action space is discrete with 4 actions which correspond to firing different actuators. The state space is continuous $\mathbb{R}^8$ and gives position and velocity information. Positive reward is given for safely landing and negative reward is given for firing the actuators and crashing.

We generate different policies with differing performance levels to mimic access to policy information from multiple human pilots. We create a constraint set of 10 constraints each of which limits the available action to that of the policy of one of the past policies or a mixture of them. 
\begin{figure*}[t]
     \centering
     \begin{subfigure}[b]{0.85\textwidth}
         \centering
         \includegraphics[width=\textwidth]{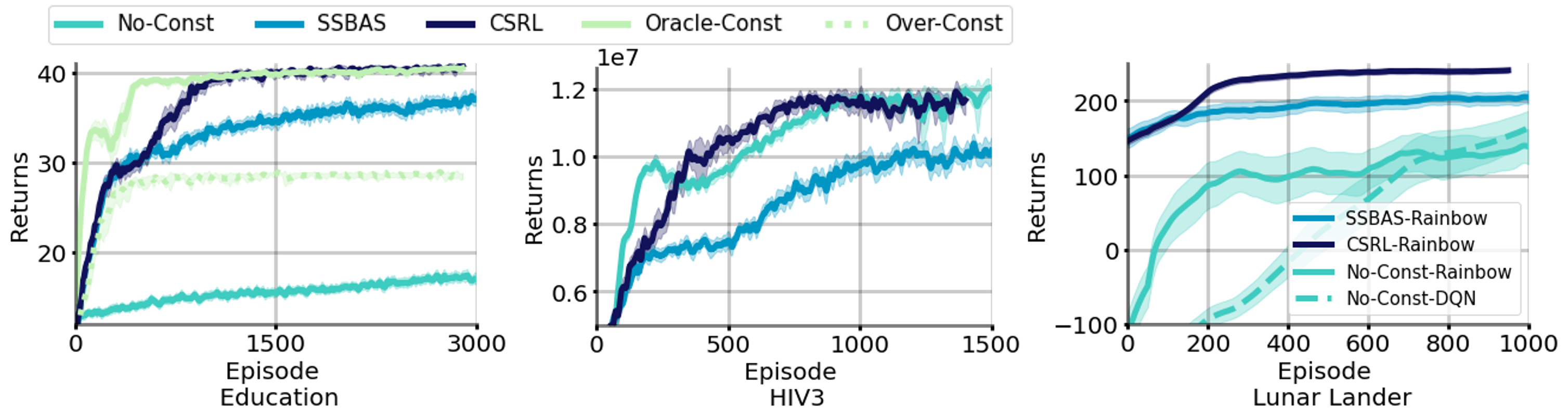}
         \caption{Training Returns}
         \label{fig:comb_returns}
     \end{subfigure}
     \begin{subfigure}[b]{0.3\textwidth}
         \centering
         \includegraphics[width=\textwidth]{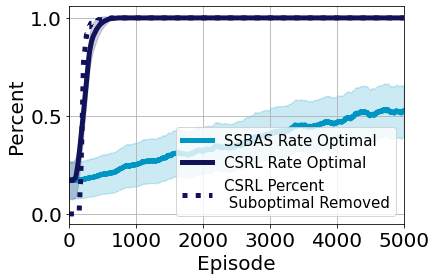}
         \caption{Lunar: Rate of optimal}
         \label{fig:rates}
     \end{subfigure}
     \begin{subfigure}[b]{0.3\textwidth}
         \centering
         \includegraphics[width=\textwidth]{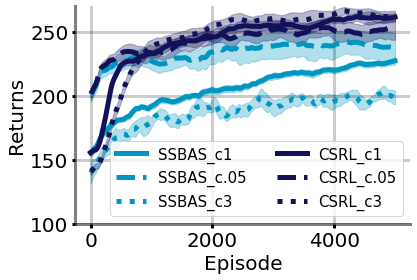}
         \caption{Lunar: c sweep}
         \label{fig:csweep}
     \end{subfigure}
     \begin{subfigure}[b]{0.3\textwidth}
         \centering
         \includegraphics[width=\textwidth]{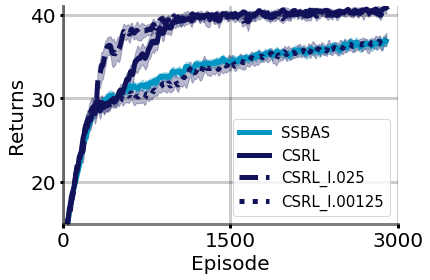}
         \caption{Education: $T_l$ sweep}
         \label{fig:lsweep}
     \end{subfigure}
     
\caption{We plot all values with 95\% confidence intervals. \textbf{Top Row} We plot episode returns during training from the Education, Lunar Lander and HIV domains. We plot our algorithm CSRL, as well as the SSBAS~\cite{laroche2018reinforcement} and unconstrained (labelled No-Const) baselines. In the educational domain, a misspecified constraint, labelled Over-Const, is also shown. The Oracle-Const plots the performance of following the oracle best constraint in the set which is not known beforehand.
In the \textbf{bottom row}, we provide further experiments and visualizations used to illustrate various properties of our algorithm for our discussion ( Section~\ref{sec:results_discussion}). Additional plots including parameter sweeps in other environments and plots over longer episodes are provided in the Appendix.
}
        \label{fig:results}
\end{figure*}

\subsection{Results and Discussion}\label{sec:results_discussion}

We compare CSRL against 4 baselines: unrestricted reinforcement learning, reinforcement learning with the \textit{oracle-constraint} which is the best constraint (but unknown in real situations), reinforcement learning under a non-oracle constraint, and SSBAS~\cite{laroche2018reinforcement}, a prior algorithm for selecting across learners with different hyperparameters or models that uses a UCB selection approach but does not eliminate learners.

We used UCRL as a base learner for our Recommendation System experiments, DDQN for Education and HIV Treatment, and Rainbow for Lunar Lander. In each experiment, the same base learning model parameters (model architecture, learning rates, exploration rate scheduling, etc) were used for all algorithms: see appendix for details.  %
For CSRL and SSBAS, total episode return was scaled to the range $[0, 1]$ for the adaptive learner selection and we used the confidence bound $B(h,s) = c \frac{\sqrt{\log(t)}}{s^{1/2}}$. This bound  satisfies the conditions of $B(h,s)$ required for Theorem 1 to hold. We did not tune $c$ and used $c = 1$ for both CSRL and SSBAS for all experiments. We later present a discussion of sensitivity to different values of $c$. For CSRL, $T_n$ was set to $20$ and $T_l$ was set to $0.05$ 
for all experiments. We did not tune either parameter and we later present a discussion of robustness to different values of $T_l$. For each experiment, the results were averaged over multiple random seeds, with 200, 20, 20, and 40 seeds for the recommendations, education, HIV treatment, and Lunar Lander experiments respectively.
\begin{table}[t]
  \fontsize{9}{9}\selectfont
  \centering
\begin{tabular}{c|ll|ll}
    \toprule
    &\multicolumn{2}{c|}{90\% Of Max}&\multicolumn{2}{c}{97\% Of Max}\\
    &SSBAS & Unconst &  SSBAS & Unconst \\
    \midrule
    Edu & 3.0 $\pm$ 0.3 & 20+  & 4.54 $\pm$ 0.6 & 20+\\
    Mov. & 1.0 $\pm$ 0.07 & 1.5 $\pm$ 0.1 & 1.7 $\pm$ 0.3 & 2.1 $\pm$ 0.2\\
    HIV & 27 $\pm$ 0.8 & 5.5 $\pm$2.4 & 20+ & 15.3$\pm$ 0.7 \\
    LL &  5.2$\pm$ 0.6 &  2.7$\pm$ 0.5 & 3.4$\pm$ 0.7 & 1.6 $\pm$  0.3 \\
    \bottomrule
    
  \end{tabular}
  \caption{The sample complexity speedup of CSRL over baselines in terms of the  factor of episodes more required by baselines to achieve returns 90\% and 97\% of the maximum observed value compared to CSRL.}\label{tab:sampcompspeed}
\end{table}

\textbf{Results} Across all domains CSRL learned a good policy faster than non-oracle baselines, and often significantly faster. In Table~\ref{tab:sampcompspeed} we list the speedup of CSRL over the SSBAS and Unrestrained (Unconst) baselines in terms of the factor of episodes more compared to CSRL needed by the baselines to achieve mean returns 90\% and 97\% of the observed maximum return. In most cases, we observe CSRL can achieve a high level of performance significantly faster than baselines, often at least 1.5 times as fast and occasionally much more. In the appendix we give a table of the raw number of episodes needed to reach these performances for each environment.
In Figure~\ref{fig:comb_returns} we plot the returns through training along with 95\% confidence intervals for the Education, HIV treatment and Lunar Lander domains. These plots for the other domains are presented in the appendix. We observe CSRL is able to utilize the constraints to avoid the extended initial poor performance compared to the unconstrained condition. Additionally the elimination mechanism allows the algorithm to eliminate suboptimal learners to quickly achieve near-oracle performance. We investigate adaptive selection and elimination in depth in the discussion:

\subsection{Discussion}
\textbf{Importance of Adaptive Selection} It is expected that following a single good constraint can lead to quick learning by reducing the policy space that needs to be explored. We see this is indeed true as the oracle-constraint performs best for most cases. 
However the best constraint is not known beforehand, and following a single constraint that is misspecified can lead to suboptimal performance. We demonstrate this in the education domain, shown in Figure~\ref{fig:comb_returns} where we additionally plot the tightest restriction in the set, labeled \textit{Over-Const} (which stands for Over-Constrained). Over-Const also quickly converges but to a return much lower than optimal. We see the adaptive selection mechanism of CSRL and SSBAS over the set can leverage the constraints to learn faster than the unrestricted baseline while avoiding this potential misspecification. Additionally the benefit of adaptive selection has over standard unrestricted RL increases with action space size as larger action spaces are harder to explore. This is illustrated by comparing the speed of learning unrestrained in our Education ($|\mathcal{A}| = 51$) and Lunar Lander ($|\mathcal{A}|=4$) environments (Fig~\ref{fig:comb_returns}).

\textbf{Importance of Elimination} In our setting where we expect some constraints to be misspecified, we found eliminating suboptimal learners to be very important for robustness against performance decreases due to missspecification. This is illustrated in the HIV experiments shown in Figure~\ref{fig:comb_returns}. In this case, all constraints are suboptimal and the unconstrained option is optimal. We notice that CSRL is able to quickly use elimination to approach the unconstrained performance compared to SSBAS.
In Figure~\ref{fig:rates} we plot the rate of selecting the optimal constraint in the lunar lander experiment for CSRL and SSBAS through learning. We see that elimination allows the algorithm to achieve high rates of selecting the optimal policy much faster. 

\textbf{When is elimination not important?} We expect elimination to not be important when the performance gap between the optimal and suboptimal constraints is large. Intuitively a large difference is easier for the UCB mechanism to distinguish the best item so CSRL and the SSBAS baseline learn quickly to choose the best item and achieve nearly identical performance. We demonstrate a case of this in the appendix. 

\textbf{The confidence bound parameter}
In Figure~\ref{fig:csweep}, we plot performance for various values of $c$, the multiplier on the confidence bound of the UCB constraint selection mechanism in the lunar lander environment. For both algorithms, we see that a low value, $c = 0.05$, results in higher initial performance but a much slower rate of learning which is a poor outcome for both algorithms. On the other hand, a higher value, $c = 3$ leads to significantly worse performance for SSBAS while CSRL's elimination mechanism protects the performance from suffering. The uncertainty over the value of the multiplier naturally comes in when the maximum reward is unknown. It is undesirable to underestimate the maximum reward as it may lead to a slow rate of learning. In these cases CSRL along with a high estimate of maximum reward can lead to good performance.

\textbf{The effect of the loss threshold $T_l$}: In Figure~\ref{fig:lsweep} we plot the performance for various values of $T_l$, the model change threshold for the elimination mechanism, in the education environment. We demonstrate that even for $T_l$ set to large values ($T_l = 0.25$, 5 times greater than initial used value), the performance does not decrease (in fact it increases due to faster elimination). This shows our robust elimination procedure is able to maintain performance when the algorithm incorrectly hypothesizes a constraint as suboptimal. When $T_l$ is set to low values ($T_l = 0.00125$, 40 times less than initial), the elimination of constraints slow and we approach the performance of the SSBAS baseline.

\textbf{Summary} Overall when there is a good constraint in the set, we demonstrate our algorithm is (1) able to achieve a good policy quickly, often significantly faster than baselines (2) this performance improvement is due to CSRL's robustness against both  misspecified constraints and hyperparameters such as the confidence bound parameter.

\section{Related Work}
We discuss some additional areas of related work not previously mentioned.

\textbf{Constrained RL} Our work is related to work in constrained RL. Most prior work considers learning under a single constraint that is always enforced. Constrained RL has been studied under various combinations of what is known and unknown about the components of the model (the constraints, rewards, and transition dynamics). It has been studied in cases where all components are known ~\cite{altman1999constrained}, as well as cases where one or more of them need to be learned online~\cite{efroni2020exploration,zheng2020constrained,achiam2017constrained,wachi2020safe,bhatnagar2012online}. This work spans a variety of different algorithms including tabular~\cite{efroni2020exploration}, policy search~\cite{achiam2017constrained}, and actor-critic~\cite{bhatnagar2012online} RL algorithms. Contrary to this work that focuses on learning to satisfy a single constraint, we consider a set of weak constraints, which may or may not be compatible with the action selections of the unknown optimal policy. 
We additionally note our method is separate  from the constraint sampling algorithms for  solving unconstrained dynamic programming  problems approximately using linear programming~\cite{de2004constraint}. 

\textbf{Inferring Constraints/Rules} 
There has been prior work on inferring constraints or rules from demonstrations~\cite{taylor2011integrating}.
In two papers
~\cite{noothigattu2019teaching,balakrishnan2019using}  a single constraint is inferred from demonstrations, and a 2-armed bandit learns if the inferred constraint should be followed. Our work differs in that we consider utilizing domain knowledge instead of demonstrations and we consider multiple potential constraints as opposed to a single constraint. We additionally differ in considering a method for RL learner elimination.

\section{Conclusion}
There often exists domain expertise for real world systems that can be leveraged in RL algorithms to speed learning. In this work we propose a method, CSRL, to incorporate this knowledge in the form of constraints. As it is often difficult to create a single constraint the algorithm designer is confident is correct, our work takes as input a set of potential constraints the algorithm designer hypothesizes, but does not have to be certain will speed learning. We provide a brief theoretical discussion on our upper confidence with elimination selection algorithm and focus on showing strong empirical results. 
We show this simple approach is compatible with deep RL methods and that CSRL can learn a good policy substantially faster than state-of-the-art baselines,  suggesting its potential for increasing the range of applications with RL is feasible by leveraging imperfect human guidance. 

\section{Acknowledgements}
This material is based upon work supported by the Stanford Human Centered AI Hoffman Yee grant and the Graduate Fellowships for STEM Diversity.

\bibliography{aaai22}

\clearpage
\newpage 
\appendix
\section*{Appendix}
\section{Additional Results}
\begin{figure*}[t]
     \centering
     \begin{subfigure}[b]{0.3\textwidth}
         \centering
         \includegraphics[width=\textwidth]{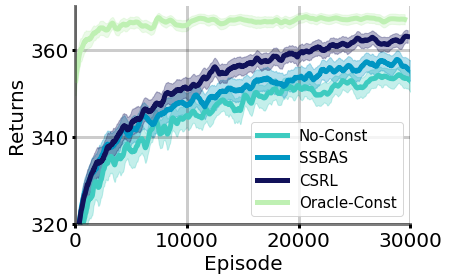}
         \caption{Movie Recs: Training Returns}
         \label{fig:mres}
     \end{subfigure}
     \begin{subfigure}[b]{0.3\textwidth}
         \centering
         \includegraphics[width=\textwidth]{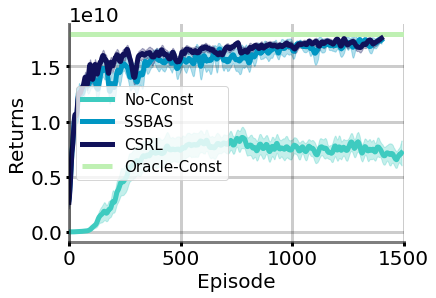}
         \caption{HIV2: Training Returns}
         \label{fig:hivres}
     \end{subfigure}
     \begin{subfigure}[b]{0.3\textwidth}
         \centering
         \includegraphics[width=\textwidth]{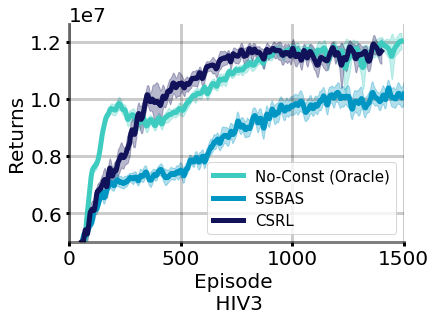}
         \caption{HIV3: Training Returns}
         \label{fig:hiv2res}
     \end{subfigure}
     \begin{subfigure}[b]{0.3\textwidth}
         \centering
         \includegraphics[width=\textwidth]{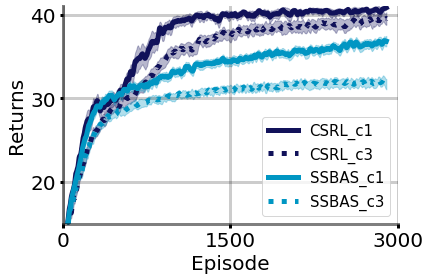}
         \caption{Edu.: Confidence Bound (c) sweep}
         \label{fig:ecsweep}
     \end{subfigure}
     \begin{subfigure}[b]{0.3\textwidth}
         \centering
         \includegraphics[width=\textwidth]{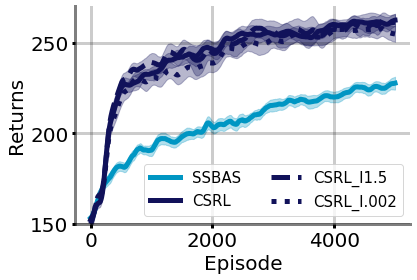}
         \caption{Lunar: $T_l$ sweep}
         \label{fig:elsweep}
     \end{subfigure}
     \begin{subfigure}[b]{0.3\textwidth}
         \centering
         \includegraphics[width=\textwidth]{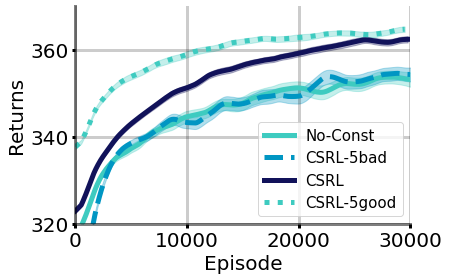}
         \caption{Movie: Smaller Constraint Sets}
         \label{fig:diff_RS}
     \end{subfigure}
     \begin{subfigure}[b]{0.3\textwidth}
         \centering
         \includegraphics[width=\textwidth]{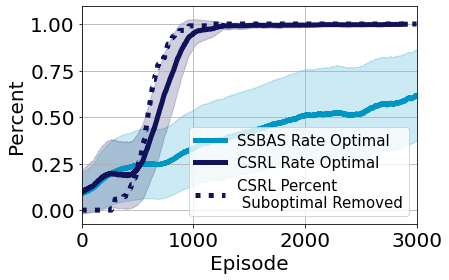}
         \caption{Edu: Optimatl Constaint Rates}
         \label{fig:ecrates}
     \end{subfigure}
     \begin{subfigure}[b]{0.3\textwidth}
         \centering
         \includegraphics[width=\textwidth]{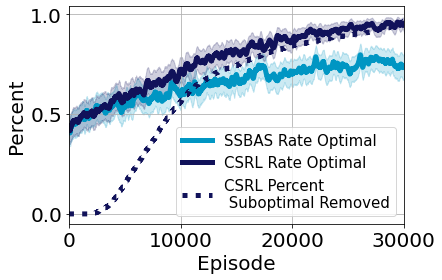}
         \caption{Movie: Optimal Constraint Rates}
         \label{fig:mcrates}
     \end{subfigure}
     \begin{subfigure}[b]{0.3\textwidth}
         \centering
         \includegraphics[width=\textwidth]{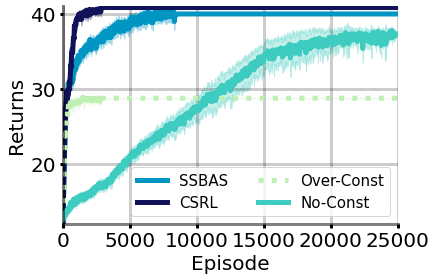}
         \caption{Edu: Longer Run}
         \label{fig:edulong}
     \end{subfigure}
     
\caption{Additional results for the results and discussion sections in domains not given in the main text. \textbf{Top Row}: Additional curves for episode returns during training. \textbf{Middle Row} Additional plots for sweeps of the confidence bound, $c$ and elimination, $T_l$, hyperparameters as well as the effect of different constraint sets. \textbf{Bottom Row} Additional plots demonstrating the rates of taking the optimal constraint for the adaptive selection algorithms as well as a longer run in the education domain.}
        \label{fig:appresults}
\end{figure*}

We provide the table listing the number of episodes required for various methods to achieve average returns 90\% and 97\% of the maximum observed return of the environment in Table~\ref{tab:sampcomp}. We report this for our method as well the SSBAS and unconstrained baselines, and the oracle best constraint which is not known beforehand. CSRL can achieve good policies significantly faster than baselines. We provide plots related to the results and discussion of the main text for domains not given in Figure~\ref{fig:appresults}. In the top row, we give training returns for the Movie Recommendations and two additional HIV treatment domains. We show in our Movie Recommendation domain our algorithm is able to perform better than baselines. 

The two HIV domains were designed to provide discussion. The second HIV treatment domain, HIV2, was constructed to demonstrate a case where elimination did not matter as mentioned in the main text. In this domain, the optimal constraint resulted in returns much higher than suboptimal returns, allowing the UCB algorithm to quickly choose the best constraint even without elimination. The third HIV treatment domain, HIV3, was constructed to demonstrate a case where all constraints were suboptimal and the unconstrained option was the best item. We demonstrate that elimination is necessary for faster performance and CSRL is able to quickly eliminate the constraints and achieve good performance.

In Figures~\ref{fig:ecsweep} and~\ref{fig:elsweep} we demonstrate the performance of different confidence bounds parameters $c$ and elimination parameters $T_l$. As in the main text we notice CSRL is able to recover better from misspecified values of $c$ and is robust to most selections of $T_l$.
In Figure~\ref{fig:diff_RS} we examine sampling over smaller constraint sets. Compared to using a smaller set, using a larger set can decrease performance. In the figure we see sampling over a set of 5 constraints which contains the optimal constraint (CSRL-5good) leads to better performance than sampling over the original set of 13 (CSRL). However if the constraint set is chosen incorrectly and does not include the best constraint then performance is worse (CSRL-5bad). This demonstrates the trade-off in using a smaller constraint set.

In the last row we plot the rate the optimal constraint is taken for SSBAS and CSRL as well as the rate suboptimal constraints are removed for the Education and Movie Recommendation domains. We notice as we saw with Lunar Lander in the main text, CSRL quickly learns to take an optimal constraint.

\section{Environments and Experimental Details}
\subsection{Recommendation System}
\textbf{Further Environment Details}
As mentioned, our recommendation system environment is a slightly modified version of the environment introduced by Warlop et al.~\cite{warlop2018fighting}. It considers the problem of movie recommendations and was created using the Movielens 100K dataset. It has a finite action set $A$ where each action is a genre of movie. Note that grouping by genre is just one way to create a manageable number of actions out of a large amount of movies, and other groupings would also work. The state space at time $t$, $s_t$, is the sequence of the last $w$ actions (formally, $s_t = (a_{t-1}, ..., a_{t-w})$). The rewards correspond to the ratings of the recommended movies and is described as a linear model. Let the following function of state and action be denoted as the \textit{recency} function: 
$$\rho(s_t, a) = \sum_{i = 1}^{w} \frac{\mathbbm{1}\{a_{t-i} = a\}}{i}$$
Intuitively, this captures how often the user has experienced the action in recent history, as $\rho(s_t, a)$ decreases if $a$ was seen less in the last $w$ movies. We next define a \textit{variability} function, $v(s_t, a)$, as the number of unique actions in the union of $s_t$ and $a$ (formally, $v(s_t,a) = \sum_{a' \in A} \mathbbm{1}\{a' \in (s_t \cup a)\}$). This metric is a proxy for the factors of diversity and novelty. Note that the number of unique values the variability function can take on is $\min(w+1, |A|)$ which is often smaller than the state space (where $|S| = |A|^w$). 
When action $a$ is taken in state $s_t$, the model assumes the observed reward is generated by a degree $d_\rho$ polynomial of $\rho$ and a degree $d_v$ polynomial of variability and a degree $d_v$ polynomial of the variability-recency cross-term. Formally, define
\begin{align*}
    x_{s,a} = [&1, \rho(s_t, a), ..., \rho(s_t, a)^{d_\rho}, v(s_t, a),.. ,v(s_t, a)^{d_v}, \\
    &v(s_t, a)\rho(s_t, a), ... (v(s_t, a)\rho(s_t, a))^{d_v}] \in \mathbb{R}^{d}
\end{align*} 
where $d = d_\rho + 2d_v +1$, then the reward $r$ is defined by:
    $$r(s_t, a) = \sum_{j = 0}^{d} \theta^*_{a,j} \rho(s_t, a)^j = x_{s, a}^T \theta^*_{a}$$
where the parameters $\theta^*$ are fit using the ratings data of the Movielens 100K dataset.
The transition model transitions to a terminal state with a probability dependent on action and variability. We fit the probability of termination for every (variability $v$, action $a$) pair. If it does not terminate, it executes a deterministic transition naturally defined by the state space. A visualization of this is given in Figure~\ref{fig:rec_sys_visual}. In our experiments, we used an action space of size $5$, $w = 4$, $d_\rho = 5$, and $d_v = 2$.

\begin{table*}[t]
  
  \fontsize{9}{9}\selectfont
  \centering

\begin{tabular}{c|llll}
  
    \toprule
    
    &CSRL & SSBAS &  Unconst & Oracle \\
    \midrule
    \multicolumn{5}{c}{90\% Of Max}\\
    \midrule
    Edu & 790 $\pm$ 42 & 2410 $\pm$ 103  & 16K $\pm$ 1K & 360 $\pm$ 33\\
    Mov. & 1408 $\pm$ 53 & 1402 $\pm$ 59 & 2160 $\pm$ 117 & 0 $\pm$ 0 \\
    HIV & 288 $\pm$ 9 & 7875\textsuperscript{*}  & 1600 $\pm$ 650 & 0 $\pm$ 0\\
    LL &  580$\pm$ 41 &  3011$\pm$ 190 & 1600$\pm$ 168 & 201 $\pm$  83 \\
    \midrule
    \multicolumn{5}{c}{97\% Of Max}\\
    \midrule
    Edu & 1144 $\pm$ 106 & 5200 $\pm$ 310  & 30K+ &932 $\pm$ 144\\
    Mov & 14.4K $\pm$ 1.2K & 25.1K $\pm$ 2.6K & 30K+ & 0 $\pm$ 0\\
    HIV & 940 $\pm$ 280 & 15K+ &  11K$\pm$9.7K & 0 $\pm$ 0 \\
    LL &  1330$\pm$ 138 & 620 &  2145$\pm$ 180 & 1273$\pm$ 240 \\
    \bottomrule
  \end{tabular}
  \caption{Sample Complexity: The table gives number of episodes necessary to learn a policy that achieves a mean return of 90\% and 97\% of the observed maximum return. CSRL can often learn a good policy many times faster than baselines.}\label{tab:sampcomp}
\end{table*}

\textbf{Modifications} In the work that introduced the environment, Warlop et al. consider the infinite horizon case with the assumption users utilize the system for infinite time whereas in this work we consider the episodic setting and consider user dropout.  We augment their environment by defining the variability function and fitting a model of dropout as well as slightly augmenting the rewards model with variability. In their original environment, Warlop et al. used a reward model that was only defined with recency: 
$$x_{s,a} = [1, \rho(s_t, a), ..., \rho(s_t, a)^{d_\rho}] \in \mathbb{R}^{d_\rho + 1}$$
Warlop et al. found this linear model of reward was able to fit the data well and we found our augmented model gave a slight, but not significant, better fit.

We additionally include a termination model. The termination model was inspired by prior work that suggests diversity and novelty is correlated with system usage and to further motivate the model, we found a statistically significant relationship between variability and dropout.  We
estimate values of dropout probability at every $v \in (2, 3, 4, 5)$, which were $(0.014, 0.0117, 0.0113, 0.0102)$ respectively. In general dropout probability decreases as variability increases, potentially indicating experiencing a wide variety of movies is important for users. Using a chi-squared test of homogeneity, we found a statistically significant difference between $v = 2$ and $v = 5$ ($\chi^2 = 3.9, p = 0.047$). The parameters we used for $\theta^*$ and $p(v,a)$ is given in the code.

\textbf{Constraints}
We include constraints on variability as defined by our variability function as defined by our variability function $v(s_t, a)$. We include constraints that require the action taken satisfies a minimum variability, which we will refer to as 'l\texttt{\#}' where \texttt{\#} (ex. 'l3' means at least 3 variability). We also include constraints that require an exact level of variability (referred to as 'e\texttt{\#}') which, if correct would induce a smaller policy space than the 'l' constraint and consequently faster learning. Additionally we introduce constraints that require an exact level of variability and do not allow the least popular action (which we preface with 'o') or the two least popular actions (prefaced with 't' ). We also include the unconstrained option (labelled as 'u').

\textbf{Hyperparameters and Experiment details}
We follow a slightly modified UCRL algorithm for estimating confidence intervals for our model parameters (the reward values $r_{s,a}$ and termination probabilities $p_{s,a}$) for efficiency. Because our domain's reward is generated from a linear model, we use the efficient linUCRL from Warlop et. al~\cite{warlop2018fighting} algorithm which leverages the linear reward structure to estimate confidence intervals for the reward parameters. Additionally, in our domain, we do not need to learn all the transition probabilities between every state-action pair for the transition model. We additionally increase efficiency by leveraging the transition model structure and only estimating the probabilities necessary, which are the termination probabilities for each variability-action pair.
\begin{figure}[t]
\begin{minipage}{0.46\textwidth}
    \begin{algorithm}[H]
    \SetAlgoLined
     \textbf{Inputs:}Policy, $\pi$\\
     Episode horizon length, $H$ \\
     Current timestep in episode, $t$\\
     Max problem limit, $l$\\
     Desired Proficiency level, $m_t$\\
     \textbf{Initialize:}
     $J = 0$, returns\\
     $\mathcal{P} = [ \ ]$, Problem Sequence so far\\ 
     $\mathcal{C} = [ \ ]$, Answer Sequence so far\\ 
     $S = [0]*(|\mathcal{A}| + \log_2(H))$\\
     $t = 0$
     
     \While{$t < H$}{
      $a \leftarrow \pi(S)$\\
      $m'_a = S[a]$\\
      $m_a$, $\mathcal{P}$, $\mathcal{C}$, $t$, $r$ = ST($a$, $\mathcal{P}$, $\mathcal{C}$, $t$, $m'_a$)\\
      $J \leftarrow J + r$\\
      $S[a] = m_a$\\
      $S[-\log_2(H):end] = binary(t)$
     }
     \textbf{Return:} $J$
     \caption{Run Episode DKT Env.}
     \label{alg:student_pseudocode}
\end{algorithm}
\end{minipage}
\hfill
    \begin{minipage}{0.46\textwidth}
    \begin{algorithm}[H]
    \SetAlgoLined
     \textbf{Input:} Action (problem), $a$ \\
     Problem Sequence so far, $\mathcal{P}$\\ 
     Answer Sequence so far, $\mathcal{C}$\\ 
     Current timestep in episode, $t$\\
     Previous Proficiency on $a$, $m'_a$\\
    \textbf{Environment Vars:} $H$, $l$, $m_t$\\
     $m_a \leftarrow$  DKT($\mathcal{P}$, $\mathcal{C}$, $p$)\;
     $n = 0$\\
     \While{$m_a < m_t$ and $n < l$ and $t < H$}{
      $\mathcal{P} \leftarrow [\mathcal{P}, a]$, $\mathcal{C} \leftarrow [\mathcal{C}, 1]$ \\
      $m_a \leftarrow$  DKT($\mathcal{P}$, $\mathcal{C}$, $a$)\\
      $t \leftarrow t + 1$, $n \leftarrow n + 1$\\
     }
     $r = (m'a < m_t) \textrm{ and } m_a > m_t$\\
     \textbf{Return:} $m_a$, $\mathcal{P}$, $\mathcal{C}$, $t$, $r$
     \caption{StudentTransition (ST)}
     \label{alg:student_transition}
\end{algorithm}
\end{minipage}
\end{figure}
All experiments were run with on a cluster machine with 2CPUs and 2GB of memory. 
For robustness we additionally add a slight tolerance to the elimination mechanism of CSRL and only eliminate an item if the mean return estimate is 2 standard deviations below the best return estimate. The full constraint structure is given in Table~\ref{tab:constraint_structure} and a visualization of the constraint structure is given in Figure~\ref{fig:const_visual}. 

\subsection{Educational Sequencing}
\begin{figure*}[t]
     \centering
     \begin{subfigure}[b]{0.25\textwidth}
         \centering
         \includegraphics[width=\textwidth]{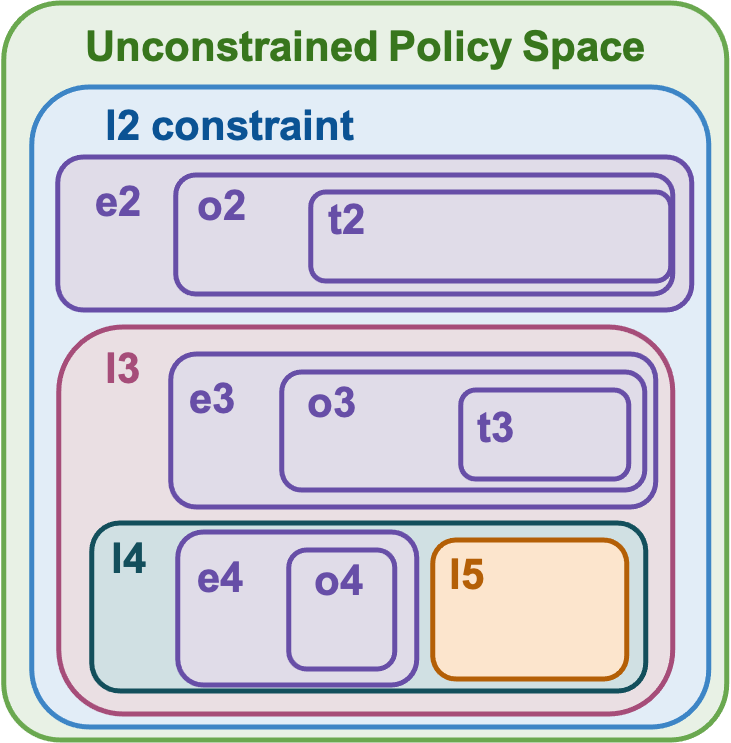}
         \caption{Visualization}
         \label{fig:const_visual}
     \end{subfigure}
     \centering
     \begin{subfigure}[b]{0.39\textwidth}
         \centering
         \includegraphics[width=\textwidth]{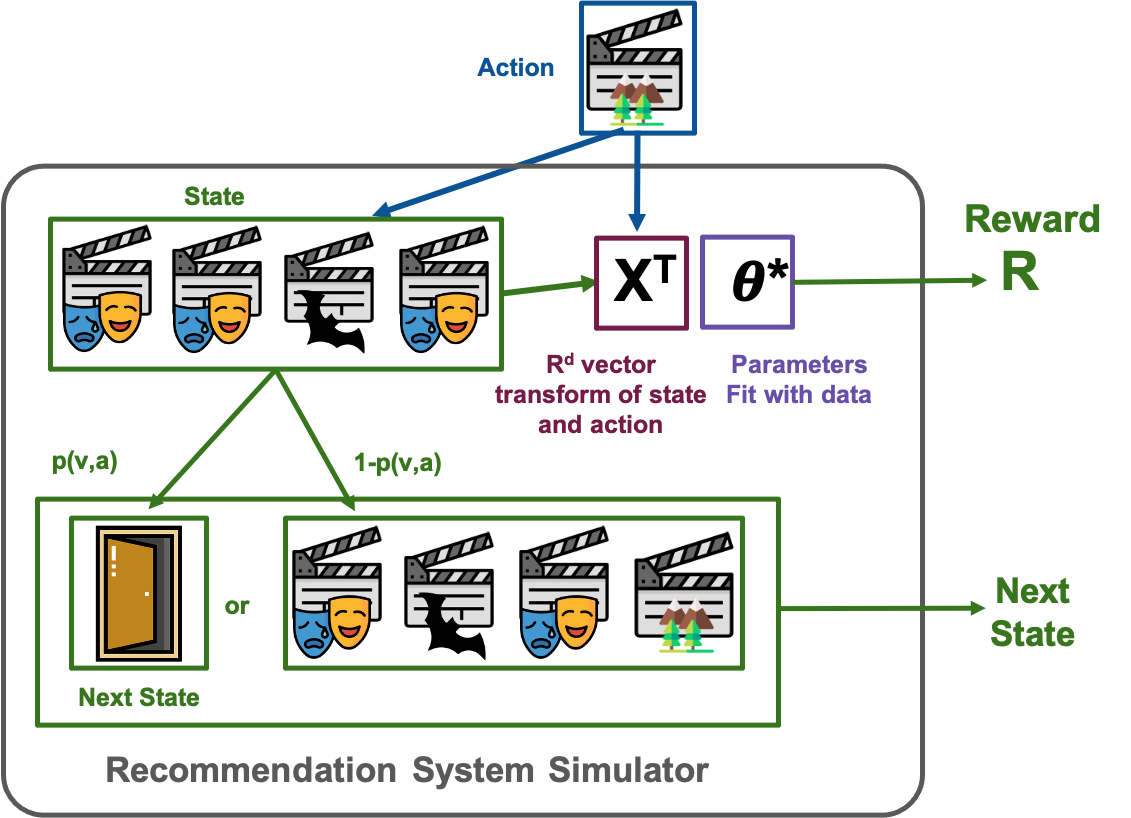}
         \caption{Rec. System Simulator}
         \label{fig:rec_sys_visual}
     \end{subfigure}
     \begin{subfigure}[b]{0.32\textwidth}
         \centering
         \includegraphics[width=\textwidth]{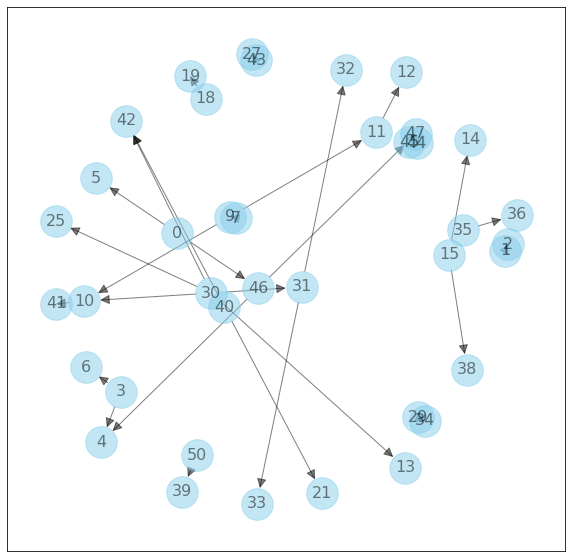}
         \caption{c30}
         \label{fig:c30}
     \end{subfigure}
\caption{(a) A visualization of the complete constraint structure for Recommendation System experiments. The image visualizes which policies are allowed under different constraints. For example all policies allowed under the 'l3' constraint are also allowed under the 'l2' constraint. Similarly all policies allowed under the 'e2' constraint are allowed under the 'l2' constraint. However the policies allowed under the 'e2' and 'l3' constraints do not overlap. (b) A visual representation picturing the recommendation system simulator. (c) The prerequisite graph corresponding to c30 in the education environment which has 30 edges. }
        \label{fig:images}
\end{figure*}
\textbf{Further Environment Details}
As mentioned in the main text, introduce a new educational activities sequencing environment which simulates a student using the popular Deep Knowledge Tracing (DKT) model~\cite{piech2015deep} of student learning trained on the ASSISTment 2009-2010~\cite{feng2009addressing} dataset. 
The model takes as input a sequence of problems attempted by the student so far and a corresponding sequence of binary indicators of answer correctness and uses a Recurrent Neural Netowrk (RNN) to predict student answer correctness probability for all the problems in the problem bank if they are presented next.
The action space of the environment is discrete of size 51 where each action corresponds to different problem being presented to the student. Mimicking the case where students have a fixed amount of time, such as a classroom setting, the horizon is fixed of length $H = 100$. When an action is taken, we input the problems presented to the student so far and observe the output predicted probability of the selected problem, which we will refer to as the \textit{proficiency} of the problem. The state space is continuous of size $|\mathcal{A}| + log_2 H$ and encodes the last observed proficiencies on all the problem as well as the current timestep. A reward rewards for mastery and is $1$ when the proficiency of a problem first exceeds a threshold $m_t = 0.85$ and $0$ otherwise.

Some additional points about the environment we would like to note is while DKT outputs the proficiency of all the problems in the problem bank, when we take an action we assume we can only observe the output corresponding to the action (equivalently, the problem) taken. This is to mimic the real world where we cannot observe the internal mental state of a student on all problems, only their performance when they attempt a problem. We also note that the answer sequence corresponding to the problem sequence we input to the DKT is always 'correct'. This is to mimic learning and assumes the student uses all resources available such as textbooks and assistance to learn to solve the problem correctly. Lastly we note that the actions can be sticky and may last more than one timestep. To mimic the cost of context switching between different topics too frequently,  we have an environment parameter $l = 10$ where once a problem is given, we present that problem to the student until either the problem is mastered, the end of the horizon is reached, or the problem has been presented $l$ times. Therefore while the horizon $H$ is fixed at $100$ timesteps, a variable number of actions are taken each episode. The full pseudocode of the environment is given in Algorithm~\ref{alg:student_pseudocode} and pseudocode for a single transition is given in Algorithm~\ref{alg:student_transition}.

\textbf{Constraints}
Using the automatic generation methods outlined in Piech et al\cite{piech2015deep}, we generate 13 graphs that have between 8 edges and 100 edges which we will denote as \textit{c}\texttt{\#} with \texttt{\#} representing the number of edges (ex \textit{c8}). The full constraint structure of the constraint set used is given in Table~\ref{tab:constraint_structure} .An additional prerequisite graph, the 'c30' graph is given in Figure~\ref{fig:c30}.

\textbf{Hyperparameters and experiment details}
We ran all experiments on cluster resources with 2CPUs, 1GPU and 16GB of memory. We use a DDQN architecture in our experiments. We use a separate model for each constraint but share collected data. For computational efficiency, we implement the data sharing with a separate experience buffer for each constraint, and adding experience collected from any constraint to the buffers of each constraint the experience is allowed under. Our model has 2 layers each with 400 hidden units and relu non-linearities. We used a batch size of 32, a learning rate of 1e-4, and copied the train network to the target network every $200$ timesteps. The one hyperparameter formally investigated was the learning rate. We additionally tried learning rates of 1e-3 and 1e-5 and found the used rate to perform the best. We decayed epsilon after every training update and used epsilon decays 0.99999 for the unconstrained option,and 0.999 for all constrained options. 

\subsection{HIV Treatment}
\textbf{Constraints}
As mentioned in the main text, we consider the case where a population consists of multiple heterogenous groups (which can be created by perturbing the internal hidden parameters of the system~\cite{killian2017robust}) and the optimal policy (such as an established doctor's treatment plan) exists for some known groups. We consider using the known optimal policies as constraints to learn in a new group which may or may not be similar to a known group and create a constraint set of 7 constraints. 
We will name the oracle based constraints \textit{b}\texttt{\#} and \textit{s}\texttt{\#} respectively where \texttt{\#} is the group used in the constraint and \textit{b} refers to 'best' and \textit{s} refers to 'second best'. For example $b0$ is the constraint where only the best action of group $0$ is allowed and $s0$ allows the the two best actions of group $0$. The full constraint structure of the constraint set used is given in Table~\ref{tab:constraint_structure}.

\textbf{Hyperparameters and experiment details}
We ran all experiments on cluster resources with 2CPUs, 1GPU and 8GB of memory. We use the same DDQN architecture and experience sharing as in the Education experiments. Our model is had 2 layers with 256 and 512 hidden units respectively and relu non-linearities. We used a batch size of 32, a learning rate of 2.5e-4, and copied the train network to the target network every $200$ timesteps. Due to large rewards (on the order of 1e9) in the environment, we additionally found taking a log transform of the rewards at each timestep resulted in higher returns. This non-linear transform of the rewards changes the objective and does not directly maximize total episode returns. However we found this method still resulted in the algorithm converging to higher returns than other methods, such as using a linear shift and scale transform of rewards, or no transformation of rewards at all. We additionally are able to achieve higher returns than those in the original paper of Ernst et al.\cite{ernst2006clinical}. We decayed epsilon after every training update and used epsilon decays 0.9999 for the unconstrained option,and 0.99 for all constrained options. 

\subsection{Lunar Lander}
\textbf{Constraints}
As mentioned in the main text, we consider a case where we have the policies from a group of pilots of various degrees of skill (to mimic different human pilots). We create a constraint set of 10 constraints which which limits the available action to that of the policy of a single or a mixture of multiple pilots.
We denoted the constraints as \texttt{p\#} where \texttt{\#} gives the pilot id(s) used in the constraint. For example, \texttt{p0} represents following pilot 0's policy while \texttt{p01} represents being constraint to taking an action from either pilot 0's or pilot 1's policy.

\textbf{Hyperparameters and experiment details}
We ran all experiments on cluster resources with 2CPUs, 1GPU and 8GB of memory. We used a Rainbow architecture. Our model has 2 layers with 128 hidden units each and relu non-linearities. We used a batch size of 64, a learning rate of 1e-4, and copied the train network to the target network every $4$ timesteps. 

\begin{table*}[t]
  \caption{Constraint Structures}
  
  \centering
  \begin{tabular}{ll|ll|ll|ll}
    \toprule
    \multicolumn{2}{c|}{Rec. Sys.} &\multicolumn{2}{c|}{HIV Treat.} & \multicolumn{2}{c|}{Education}& \multicolumn{2}{c}{Lunar Lander} \\
    Const.    & Less Const & Const.    & Less Const & Const.   & Less Const & Const.    & Less Const \\
    \midrule
     U & [] & U & [] & U & [] & U & [] \\
     g2 & [U] & s0 & [U, b0] & c8 & [U] & p0 &[U, p01] \\
     g3 & [U, g2] & s1 & [U, b1]  & c30 & [U, c8] & p1 &[U, p01]  \\
     g4 & [U, g2, g3] & s2 & [U, b2]  & c50 & [U, c8, c30] & p2 &[U, p23]  \\
     g5 & [U, g2, g3, g4] & b0 & [U] & c55 & [U, c8, c30, c50] &p3 &[U, p23]  \\
     e2 & [U, g2, g3] & b1 & [U] & c60 & [U, c8, c30, c50, c55] &p4  &[U, p45]  \\
     e3 & [U, g3] & b2 & [U] & c65 & [U, c8, c30, c50, c55, c60] &p5 &[U, p45]  \\
     e4 & [U, g4] & & & c70 & [U, c8, c30, c50, c55, c60, c65] & p01 &[U] \\
     o2 & [U, g2, e2] & & & c75 & [U, c8, c30, c50, c55, c60, c65, c70] & p23 &[U]   \\
     o3 & [U, g3, e3]  & & & c80 & [U, c8, c30, c50, c55, c60, ... c70, c75] & p45 &[U]  \\
     o4 & [U, g4, e4]  & & & c85 & [U, c8, c30, c50, c55, c60, ... c75, c80]&&  \\
     t2 & [U, g2, e2, o2] & &  & c90 & [U, c8, c30, c50, c55, c60, ... c80, c85] &&  \\
     t3 & [U, g3, e3, o3]  & & & c85 & [U, c8, c30, c50, c55, c60, ... c85, c90]&& \\
     & & & & c100 & [U, c8, c30, c50, c55, c60, ... c90, c95] &&  \\

    \bottomrule
  \end{tabular}\label{tab:constraint_structure}\\
  {\raggedright Table~\ref{tab:constraint_structure} shows the constraint structure for the Recommendation System (Rec. Sys.), HIV treatment (HIV Treat.), Education and Lunar Lander environments respectively. For each constraint (Const.), it gives the list of the less constrainted constraints (Less Const). U represents the unconstrained option.}
\end{table*}

\section{Theoretical Discussion and Proofs}
\subsection{Proof of Theorem 1}
We first restate the assumptions and definitions from the main text.

\assumpmpc*

Because the algorithm parameters determine the policy, this directly implies policy convergence.

\begin{lemma}\label{lemma:policyconv}
Policy Convergence: The expectation $\mu_{k,n}$ converges to a value $\mu_k = \lim_{n \rightarrow \infty} \mathbb{E}[\bar{J}_{k,n}] $
\end{lemma}

 Let $\pi_{\mathcal{M}_k}$ denote the policy corresponding to the converged model parameters of RL learner $l_k$, 
$\mathcal{M}_k$. Let $\mathbb{E}_{s_0}[V^{\pi_{\mathcal{M}_k}}(s_0)] = \mu_k$. Let the superscript $*$ denote optimality and let $\mu^* = \max_k \mu_k$ and $\pi^*$ denote the policy that achieves $\mu^*$. Note that $\pi^*$ is defined as the policy with the highest return across all learners in the set.   Without loss of generality assume $\pi^*$ is unique. We refer to the set of constraints compatible with $\pi^*$ as the set of optimal constraints and denote this set as $\mathcal{C}^*$ with a corresponding indices set $\mathcal{K}^*$. Then require:
\assumpc*

We then can show that at least one RL learner with a constraint set compatible with the optimal policy will be chosen strictly more than all suboptimal learners assymptotically:
\thmr*

\textit{Proof} We will first show all items that are never eliminated are taken infinitely often, we will then show at least one optimal item in $\mathcal{K}^*$ is never eliminated and finally we will show the desired result: asymptotically, at least one optimal item is taken more than all suboptimal items.

We first define some additional useful notation. Let $J_{k,n}$ denote the episode return obtained when selecting learner associated with the $k^{th}$ constraint for the $n^{th}$ time. Because we consider the case of bounded rewards and horizons, note that the episode returns are also bounded such that $|J_{k,n}| \leq J_{max}$ for all $k, n$. Let rewards be bounded such that $|r| < R_{max}$ and the horizon $h$ bounded by $H$, then $J_{max} = HR_{max}$. Let the empirical average after taking the $k^{th}$ constraint $n$ times be $\bar{J}_{k,n} = \frac{1}{n}\sum_{i=1}^n J_{k,i}$. Let the expected value of $\bar{J}_{k,n}$ be $\mu_{i,n} = \mathbb{E}[\bar{J}_{k,n}]$. Similarly, let $\mathcal{M}_{k, n}$ represent the parameters of the algorithm under the $k^{th}$ constraint after it has been taken $n$ times.

We will start off by showing the following proposition that states items that are never eliminated are taken infinitely often. Let $\mathcal{K}$ denote the set of indexes of the items that are never eliminated.
\begin{lemma}
$\lim_{h \rightarrow \infty} T_{k, h} = \infty \textrm{ for all } k \in \mathcal{K}$ 
\end{lemma}
\textit{Proof By Contradiction} 
 Assumption~\ref{thm:assumpmmpc} presents a convergence property on the learning algorithm under the constraints. We will also refer to constraints as item and the set of indices as $\mathcal{K}$.
Assume some items in $\mathcal{K}$ are taken a finite number of times. Denote this set $\mathcal{K}_{fin}$ and denote the set of items taken an infinite number of times $\mathcal{K}_{\infty}$. Note that there must be at least one item in $\mathcal{K}_{\infty}$ as there are a finite amount of items in $\mathcal{K}$. 

For each item $k$ in $\mathcal{K}_{fin}$, by definition there exists a time $N$ where that item  will never be chosen again and for $h > N$ there exist constant $c_k$:
$$T_k(h) = T_k(N) = c_k \textrm{ for } k \in \mathcal{K}_{fin} \textrm{ and } h > N$$

This implies that for any item $k \in \mathcal{K}_{fin}$ and every episode $h$:
$$\bar{J}_{k,c_k} + \frac{z(h)}{c_k^\eta} \leq \bar{J}_{j,T_j(h)} + \frac{z(h)}{T_j(h)^\eta}$$
for all $h > N$ and at least one $j \in \mathcal{K}_{\infty}$

Let $j_+$ be any of the items $j$ that satisfies this at time t. Rearranging the above and setting $\frac{1}{c_k^\eta} = c_k'$:
$$\bar{J}_{k,c_k} - \bar{J}_{j_+,T_j(h)} \leq \frac{z(h)}{T_{j_+}(h)^\eta} - c_k' z(h) \ for \ h > N$$

Looking at the right-hand side, $c_k'$ is positive, both $z(h)$ and $T_{j_+}(h)$ are monotonically increasing with $h$ and $\lim_{h \rightarrow \infty} z(h) \rightarrow \infty$ and $\lim_{h \rightarrow \infty} T_{j_+}(h) \rightarrow \infty$.
This implies:
$$\lim_{h \rightarrow \infty} z(h) \left(\frac{1}{T_{j_+}(h)^\eta} - c_k'\right) \rightarrow \lim_{h \rightarrow \infty} -z(h) c_k'\rightarrow -\infty$$
Therefore:
$$\lim_{h \rightarrow \infty} \bar{J}_{k,T_{k, N}} - \bar{J}_{j_+,T_j(h)} \leq \lim_{h \rightarrow \infty} \frac{z(h)}{T_{j_+}(h)^\eta} - c_k' z(h) \rightarrow -\infty$$
Which implies:
$$\bar{J}_{k,T_{k, N}} - \mu_{j_+} \rightarrow -\infty$$ which is a contradiction because $J$ is bounded.

Next we note at least one item in $\mathcal{K}_*$ is never eliminated. Therefore if Assumption 2 holds, the there will always be one optimal constraint that is never eliminated because we never remove an item that is the loosest constraint. A policy that is allowed under a constraint is also allowed under all looser constraints.

Finally we prove the main theorem. We have previously shown that an item in the unremoved set is taken an infinite number of times, so 
$$\lim_{t \rightarrow \infty} \bar{J}_{j,T_j(h)} \rightarrow \mu_j \textrm{ for each } j \in \mathcal{K}_{\infty}$$
For suboptimal items that have been removed, then the theorem holds trivially as those items are only selected a finite amount of times.

Assume a suboptimal unremoved item $j \notin \mathcal{K}_*$ has been taken the same amount or more times than an optimal unremoved item as $\lim_{h \rightarrow \infty}$ (formally $T_j(h) \geq T_*(h)$). We will show for every suboptimal item this is satisfied for, the unremoved optimal item will have a higher upper confidence bound and will be chosen over that item, proving $T_*(h) > T_i(h)$.

We will show this by contradiction, first assume $T_j(h) \geq T_*(h)$ and $*$ is not selected so the upper confidence of the optimal item is lower than the upper confidence of the suboptimal item.
Then after convergence:
$$\bar{J}_{*,T_j*(h)} + \frac{z(h)}{T_*(h)^\eta} \leq \bar{J}_{j,T_j(h)} + \frac{z(h)}{T_j(h)^\eta}$$
taking the limit as $t \rightarrow \infty$ and rearranging
$$\mu_{*} - \mu_{j} \leq  \frac{z(h)}{T_j(h)^\eta} - \frac{z(h)}{T_*(h)^\eta}$$
Because $T_j(h) \geq T_*(h)$, $\frac{z(h)}{T_j(h)^\eta} - \frac{z(h)}{T_*(h)^\eta} \leq 0$ therefore $\mu_{*} - \mu_{j} \leq 0$ however this is a contradiction as $\mu_{*}$ is defined to be the maximal value.

\subsection{Convergence when the algorithms have guarantees on the rates of convergence}

In the case where the learning algorithms under constraints have regret guarantees, such as UCRL, we can closely follow the techniques of Shah et al.~\cite{shah2020non} to provide a concentration property. In this subsection we will first show this generally and then provide discussion for UCRL.
\subsubsection{General Formulation}
We continue to use the notation defined previously in this section and start by introducing an additional assumption that controls the concentration of the RL algorithms.
\begin{restatable}[Concentration]{assump}{assumpconc}
\label{assump:concentration}
There exists constant $0 < \eta \geq 1/2$ and invertible function $g(z)$ such that for every $z\geq 1$ and every integer $n \geq 1$, such that the following is satisfied:
\begin{align*}
&P(n\bar{J}_{k,n} - n\mu_i \geq n^{(1-\eta)}z) \leq g(z)
\textrm{ and } \\
&P(n\bar{J}_{k,n} - n\mu_i \leq -n^{(1-\eta)}z) \leq g(z)
\end{align*}
\end{restatable}

In the proof we will use Assumption~\ref{thm:assumpmmpc} (specifically Lemma~\ref{lemma:policyconv}) defined previously in the section and the newly defined Assumption~\ref{assump:concentration}. Notice that Assumption~\ref{thm:assumpc} automatically holds for the algorithms considered.

We also require additional structure on the form of our confidence bound functions $B(h,n)$. Let $z(h) = g^{-1}(f(h))$ so the form of $B(h,n)$ is now $$B(h,n) = \frac{ g^{-1}(f(h))}{n^\eta}$$ for the $g$ and $\eta$ satisfying Assumption~\ref{assump:concentration} above. Additionally choose $f(h)$ to satisfy: (i) $O(g^{-1}(f(h))^{1/\eta}) < O(h)$, (ii) $g^{-1}(f(h))$ is increasing with $h$, and (iii) $\sum_{t = 1}^{\infty} hf(h) = c_f$ for some finite constant $c_f$.

We define some additional notation. Let $J^h$ be the random episode return obtained for the $h^{th}$ episode. Let $\bar{J}^{h} = \frac{1}{h}\sum_{i=1}^h J^{i}$ be the average return of the algorithm after $h$ episodes. Recall that $\mathcal{K}_*$ represented the indexes of all items that achieves the maximum return at convergence. Let $\mathcal{K}_{sub}$ represent the set of indexes of suboptimal items which do not achieve the maximum return at convergence. Let $\Delta_i$ be the the gap between the optimal and the $i^{th}$ suboptimal item for $i \in \mathcal{K}_{sub}$, formally defined as $\Delta_i = \mu_* - \mu_i$. Let $\Delta_{min}$ denote the smallest of $\Delta_i$, formally $\Delta_{min} = \min_{i \in \mathcal{K}_{sub}}\Delta_i$.

We additionally define $\delta_{*,n}$ to represent the speed of convergence of the items in the optimal set. Let sequence $\{k^*_i\}$ represent the slowest possible convergence of the optimal set. Formally, construct sequence $\{k^*_i\}$ in the following method: consider only sampling between items from the optimal set, at each timestep choose the item index $k^* \in \mathcal{K}_*$ such that $\mu_* - J_{k^*, n_{k^*}}$ is minimized. Let sequence $\{J_{k^*, n_{k^*}}\}$ correspond to the returns of sequence $\{k^*_i\}$.

With this definition, we can modify the proof of Theorem 3 of Shah et al.~\cite{shah2020non} to show the following convergence result:
\begin{restatable}[Convergence]{thm}{thmconv}
\label{theor:conv}
\begin{align*}
    |\mathbb{E}[\J_h]& - \mu_*| \leq  |\delta_{*,h}| +\\
    &\frac{2R_{max}(|\mathcal{K}|-1)([\frac{2}{\Delta_{min}}g^{-1}(f(h))]^{1/\eta} + 2c_f + 1) }{h}
\end{align*}
\end{restatable}

Shah et al. provide convergence and concentration results for Monte Carlo tree search (MCTS) using upper confidence bound based action selection. They showed the convergence of the action values in MCTS satisfied a concentration assumption stricter than the one we present (Assumption~\ref{assump:concentration}) and uses that assumption to show their desired results. In our method, to incorporate reinforcement learning algorithms, we have a looser concentration assumption and we only show convergence. Because we only focus on showing convergence, we are able to modify the proof to use more general notation, such as replacing specific functions with a set of functions that satisfy certain conditions while still carrying out the steps. In the proof of convergence, they first show that when each item is chosen enough times, the suboptimal items will become less likely. They next give an upper bound on the expected number of times a suboptimal item is chosen and use this to prove the main result.

This theorem considers the case where data is not shared between items. Intuitively, sharing experience should make learning more efficient.
\subsubsection{UCRL example}
Consider the case of using UCRL as a base learning algorithm. Let constraints be of the form we consider in our experiments, where we do not allow certain actions to be taken in a given state. In this case, the constrained algorithm can be seen as a unconstrained UCRL algorithm with a smaller action space.

From Theorem 2 of Jaksch et al.~\cite{auer2009near} the regret bound of UCRL is:
$$P\left(n\mu_i - n\J_i \geq 34DS\sqrt{AnH\log\left(\frac{nH}{\delta}\right)}\right) \leq \delta $$
Notice for $x>5500$ and any $0 < \delta \leq 1$, $\log(\frac{x}{\delta}) < \frac{x}{\delta}^{1/5}$. (In this case $x = nH$, so if $H = 100$ then this would hold after the $550^{th}$ episode).
Then for $nH > 5500$:
$$P\left(n\mu_i - n\J_i \geq 34DS\sqrt{AnH\left(\frac{nH}{\delta}\right)^{1/5}}  \right) \leq \delta$$
Rearranging
$$P\left(n\mu_i - n\J_i \geq n^{3/5}\left( 34DS\sqrt{AH^{6/5}\delta^{-1/5}} \right) \right) \leq \delta$$
Let $z = 34DS\sqrt{AH^{6/5}\delta^{-1/5}}$, $\eta = 2/5$, and $\delta = g(z)$ then Assumption~\ref{assump:concentration} is satisfied:
$$P\left(n\mu_i - n\J_i \geq n^{1-\eta}z) \right) \leq g(z)$$

We can then solve for $g^{-1}$, choose $f(h)$ so conditions on $f(h)$ and $g^{-1}(f(h))$ are satisfied, and solve for $c_f$ for the chosen $f(h)$ to complete the desired bound. Let $c_c = 34DS\sqrt{AH^{6/5}}$
$$g(z) = g(c_c\delta^{-1/10}) =  \delta \rightarrow g^{-1}(x) = c_c x^{-1/10}$$

Choose $f(h) = c_c^{10} t^{-2.1}$, then the three conditions on $B(h,n)$ are satisfied:
\begin{enumerate}
    \item $O(g^{-1}(f(h))^{1/\eta}) = O(h^{2.1/4}) < O(h)$
    \item $g^{-1}(f(h)) = h^{2.1/10}$ is nondecreasing in $h$
    \item $\sum_{t=1}^{\infty} h f(h) = c_c^{10} \sum_{h=1}^{\infty} h^{-1.1} = c_f$ for a finite constant $c_f$ ($\approx 10.59 c_c^{10}$)
\end{enumerate}

With these choices, we have the following convergence property for UCRL:
\begin{align*}
    |\mathbb{E}[\J_h] - \mu_*| & \leq  |\delta_{*,h}| +\\ &\frac{2R_{max}(|\mathcal{K}|-1)\left(\frac{2}{\Delta_{min}}\right)^{\frac{5}{2}} h^{\frac{2.1}{4}} + 2c_f + 1) }{h}
\end{align*}

Consider the case where the unconstrained condition is in the set which is common for robustness. The first term $|\delta_{*,h}|$ is dominated by the order of the unconstrained condition which is $O(log(h))$ while sampling between constraints gives a penalty on the order of $O(h^{2.1/4})$.
\end{document}